\documentclass[lettersize,journal]{IEEEtran}
\usepackage{amsmath,amsfonts}
\usepackage{algorithmic}
\usepackage{algorithm}
\usepackage{array}
\usepackage[caption=false,font=normalsize,labelfont=sf,textfont=sf]{subfig}
\usepackage{textcomp}
\usepackage{stfloats}
\usepackage{url}
\usepackage{verbatim}
\usepackage{graphicx}
\usepackage{cite}

\usepackage{times}
\usepackage{epsfig}
\usepackage{amssymb}
\usepackage{verbatim}
\usepackage{amsthm}
\usepackage{color}
\usepackage{soul}
\usepackage{wrapfig}
\usepackage{caption}
\usepackage{booktabs}
\usepackage{multirow}

\newtheorem{theorem}{Theorem}
\newtheorem{definition}{Definition}
\newtheorem{lemma}{Lemma}
\newtheorem{corollary}{Corollary}

\begin{document}

\title{
Task Attribute Distance for Few-Shot Learning: Theoretical Analysis and Applications}

\author{Minyang~Hu,
  Hong Chang,~\IEEEmembership{Member,~IEEE,}
  Zong Guo,
  Bingpeng Ma,~\IEEEmembership{Member,~IEEE,}
  Shiguan~Shan,~\IEEEmembership{Fellow,~IEEE}
  and~Xilin~Chen,~\IEEEmembership{Fellow,~IEEE}

  \thanks{Minyang Hu and Zong Guo are with the Key Laboratory of Intelligent Information Processing of Chinese Academy of Sciences (CAS), Institute of Computing Technology, CAS, Beijing 100190, China, and also with the School of Computer Science and Technology, University of Chinese Academy of Sciences, Beijing 100049, China (e-mail: minyang.hu@vipl.ict.ac.cn; zong.guo@vipl.ict.ac.cn).}

  \thanks{Hong Chang, Shiguang Shan and Xilin Chen are with the Key Laboratory of Intelligent Information Processing of Chinese Academy of Sciences (CAS), Institute of Computing Technology, CAS, Beijing 100190, China, also with the School of Computer Science and Technology, University of Chinese Academy of Sciences, Beijing 100049, China, and also with the Peng Cheng Laboratory, Shenzhen 518055, China (e-mail: changhong@ict.ac.cn; sgshan@ict.ac.cn; xlchen@ict.ac.cn).}

  \thanks{Bingpeng Ma is with the School of Computer Science and Technology, University of Chinese Academy of Sciences, Beijing 100049, China (e-mail: bpma@ucas.ac.cn).}
}

\markboth{Journal of \LaTeX\ Class Files,~Vol.~14, No.~8, August~2021}%
{Shell \MakeLowercase{\textit{et al.}}: A Sample Article Using IEEEtran.cls for IEEE Journals}


\maketitle
\begin{abstract}
    Few-shot learning (FSL) aims to learn novel tasks with very few labeled samples by leveraging experience from \emph{related} training tasks.
    In this paper, we try to understand FSL by delving into two key questions:
    (1) How to quantify the relationship between \emph{ training} and \emph{novel} tasks?
    (2) How does the relationship affect the \emph{adaptation difficulty} on novel tasks for different models?
    To answer the two questions, we introduce Task Attribute Distance (TAD) built upon attributes as a metric to quantify the task relatedness.
    Unlike many existing metrics, TAD is model-agnostic, making it applicable to different FSL models.
    Then, we utilize TAD metric to establish a theoretical connection between task relatedness and task adaptation difficulty.
    By deriving the generalization error bound on a novel task, we discover how TAD measures the adaptation difficulty on novel tasks for FSL models.
    To validate our TAD metric and theoretical findings, we conduct experiments on three benchmarks.
    Our experimental results confirm that TAD metric effectively quantifies the task relatedness and reflects the adaptation difficulty on novel tasks for various FSL methods, even if some of them do not learn attributes explicitly or human-annotated attributes are not available.
    Finally, we present two applications of the proposed TAD metric: data augmentation and test-time intervention, which further verify its effectiveness and general applicability. 
The source code is available at https://github.com/hu-my/TaskAttributeDistance.  
\end{abstract}


\section{Introduction}
\label{sec:intro}

\IEEEPARstart{L}{earning} in human biological system exhibits remarkable efficiency in comparison to artificial systems.
For instance, only one example is enough for a child to learn a novel word \cite{carey1978cognitive}, while thousands of training samples are needed for deep learning models.
This learning efficiency comes from the past experiences accumulated by human brain.
Inspired by human learning capability, Few-Shot Learning (FSL) is proposed, which aims to learn novel tasks with very few samples by leveraging experience from \emph{related} training tasks.

\begin{figure}
    \centering
    \includegraphics[width=1.0\linewidth]{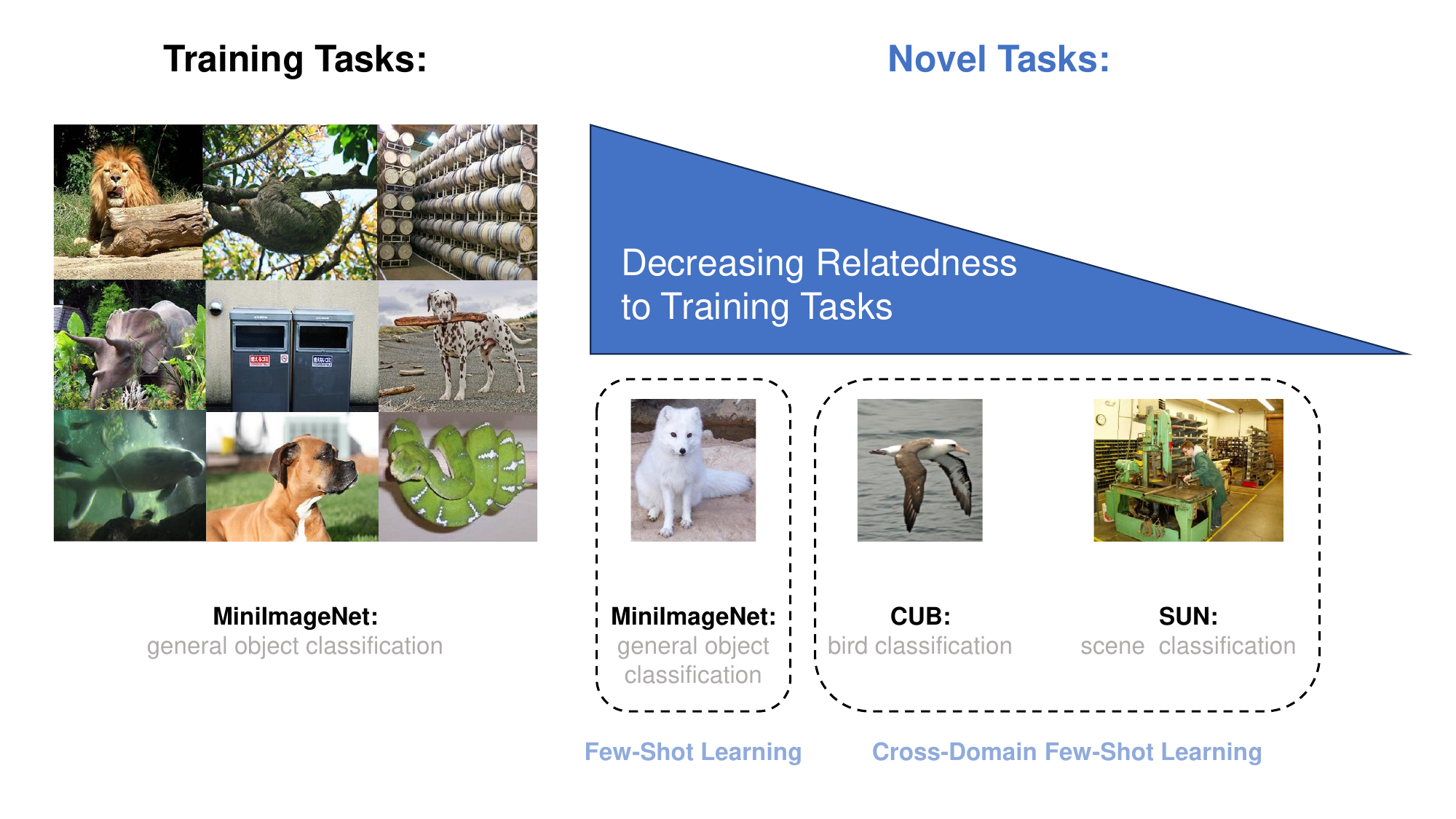}
   \caption{Different settings of Few-Shot Learning (FSL). Standard FSL focuses on constructing training and novel tasks by sampling categories within a dataset (e.g. miniImageNet). Cross-Domain Few-Shot Learning (CD-FSL) considers novel tasks sampled from different dataset (e.g. CUB or SUN).}
   \label{fig:setting}
\end{figure}

To address the FSL problem, earlier studies \cite{miller2000congealing, hariharan2017SGM, antoniou2017DAGAN, schwartz2018delta-encoder} focus on a series of \emph{data augmentation} methods.
These augmentation methods aim to learn the inter-class or intra-class variations from related training tasks and subsequently apply them to novel tasks.
These learned variations enrich the data diversity and discriminative information, which equivalently increases the number of samples.
With a sufficient number of augmented training samples in novel tasks, FSL is expected to return to the conventional supervised learning paradigm, so all supervised learning techniques can be used to solve FSL problems.
However, solving FSL problems by data augmentation comes with several weaknesses: the augmentation policy is typically tailor-made for each dataset, posing challenges in its adaptation for other datasets \cite{wang2020fsl_survey}.

A more popular solution to FSL problems is \emph{meta-learning}, which seeks to extract cross-task knowledge and facilitate the knowledge transfer from training to novel tasks by customizing the learning paradigm.
Meta-learning-based approaches in FSL area can be broadly divided into two categories. The metric-based approaches \cite{ vinyals2016matchnet, snell2017prototypical, sung2018relationnet, hou2019CAN, allen2019IMP, fei2021z-norm, afrasiyabi2022SetFeat, xie2022DeepBDC} aim to learn a cross-task embedding function and predict the query labels based on the learned distances. On the other hand, the optimization-based approaches focus on learning some cross-task optimization state, such as model initialization \cite{finn2017maml, sun2019mtl} or step sizes \cite{li2017meta-sgd, antoniou2019trainmaml}, to rapidly update models with very few labeled samples.
Despite the remarkable success achieved by meta-learning in the past years, recent findings indicate that it is effective only when the training tasks are closely related with the novel task \cite{guo2020broader, triantafillou2020meta-dataset, song2023fsl_survey23}.

To further investigate this issue, Cross-Domain Few-Shot Learning (CD-FSL) is proposed.
Different from standard FSL problem, CD-FSL considers a significant task difference (See Fig. \ref{fig:setting}).
For example, the training task involves generic object classification, whereas the novel task shifts to a more fine-grained bird classification task.
This substantial gaps in task categories inherently in CD-FSL present an obstacle to knowledge transfer, limiting the performance of model adaptation to novel tasks.
To mitigate such task difference and address the CD-FSL problem, several works propose to simulate the difference between training and novel tasks during the training phase \cite{tseng2020fwt, hu2022afa} or mine more information in novel tasks \cite{phoo2020startup, islam2021ddn, liang2021nase}.
An underlying empirical hypothesis in these works is that larger category gap leads to less relation between training and novel tasks, thereby the corresponding CD-FSL problem will be more challenging.
Although this follows the intuition, two natural questions arise for FSL and CD-FSL:
(1) How to quantify the \emph{relationship between training and novel} tasks?
(2) How does the relationship affect the \emph{adaptation difficulty} on novel tasks for different FSL models?

\begin{figure}
    \centering
    \includegraphics[width=1.0\linewidth]{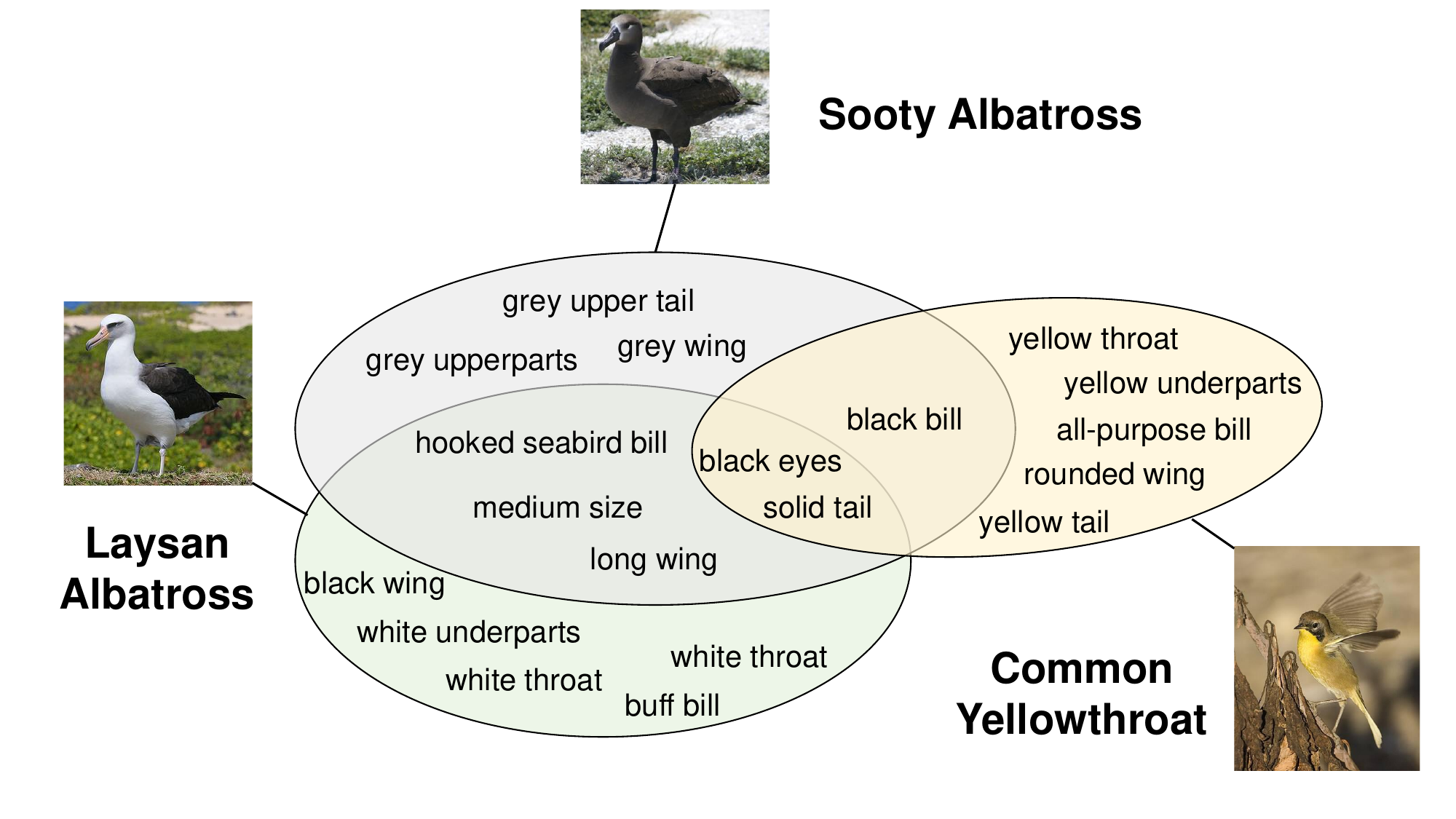}
   \caption{The categories and attribute sets of three bird images. Each category can be represented as a composition of some attributes, which act as a relationship bridge between different categories.}
   \label{fig:compositionality}
\end{figure}

To quantify the relationship between tasks, one approach is to measure the category relationship in a common representation space. 
With properly learned representation function, various distribution divergence metrics, like EMD \cite{rubner1998emd} and FID \cite{heusel201fid}, can be used to calculate the distance between feature distributions of different categories, as well as tasks over them.
The main obstacle to this solution is how to learn a good representation function for novel categories with only a few labeled samples.
Previous works \cite{milbich2021ooDML, oh2022domain-similarity} assume that the representation function learned from training categories can be directly applied to novel categories, but this assumption may not hold due to the large category gaps.
Some recent studies \cite{achille2019task2vec, le2022task-affinity}  utilize the Fisher Information Matrix to quantify the task relatedness without assuming a common representation function.
However, this approach has a heavy burden of computing Hessian matrix.
Besides, the calculated task distances are highly dependent on the learned model, making them difficult to apply to other FSL models.

To investigate the influence of task relatedness on the difficulty of adapting models to novel tasks, existing studies \cite{ren2020fsal, sariyildiz2021concept-gen, milbich2021ooDML, oh2022domain-similarity} often make an empirical assumption that a novel task with large distances to the training tasks will be hard to adapt to.
With this assumption, some works \cite{sariyildiz2021concept-gen, milbich2021ooDML, oh2022domain-similarity} employ specific distance metrics to quantify the task relatedness, and then construct more challenging benchmarks to explore the generalization ability of different models.
Different with the these works, \cite{le2022task-affinity} selects the most related training data to improve the episodic fine-tuning process for FSL models based on an asymmetric task relatedness measure.
Despite their empirical success, all the aforementioned works lack theoretical analysis, leaving the connection between task relatedness and task adaptation difficulty not formally clarified.

In this work, we try to overcome the existing obstacles and answer the two questions formally.
Firstly, we introduce \textit{Task Attribute Distance} (TAD) as a metric to quantify the task relatedness.
Our intuition lies in the \emph{attribute compositionality hypothesis} (shown in Fig. \ref{fig:compositionality}): each category can be represented as a composition of attributes, which are reusable in a huge assortment of meaningful compositions.
TAD formulates this intuition by first measuring the total variation distance between attribute distributions of training and novel categories, and then finding the maximum matching with minimum weight of a bipartite graph.
Unlike many existing metrics, TAD only relies on the attribute conditional distributions, making it independent of models and the number of samples in novel tasks.
Secondly, we utilize the TAD metric to establish a theoretical connection between task relatedness and task adaptation difficulty in FSL.
We provide theoretical proof to verify that TAD contributes to the \emph{generalization error bound on a novel task}, at different training settings.
Importantly, our theoretical results indicate that the adaptation difficulty on a novel task can be efficiently measured based on the TAD metric, without the need for training any models.

We conduct experiments on three benchmark datasets to validate our theoretical results and explore the generality of TAD metric.
To this end, we consider two scenarios that the human-annotated attributes are provided or not available.
For the latter case, we propose an efficient way to auto-annotate attributes.
Experimental results show a \emph{linear relationship} between few-shot performance and the calculated distance: as the distance between novel and training tasks increases, the performance of various FSL models on novel tasks decreases linearly. 
Based on our theoretical analysis and experimental results on benchmarks, we further present two potential applications for TAD metric for more real-world FSL settings: 
selecting the most related training tasks for data augmentation and intervening the challenging novel test tasks to improve the worst-task performance of FSL models.

Our main contributions can be summarized as follows:
\begin{itemize}
    \item We propose Task Attribute Distance (TAD) metric in FSL to quantify the relationship between training and novel tasks, which is model-agnostic and efficient to compute.
    \item We provide a theoretical proof of the generalization error bound on a novel task based on TAD, which connects task relatedness and adaptation difficulty theoretically.
    \item We conduct experiments to show TAD can effectively quantify the task relatedness and reflect the adaptation difficulty on novel tasks for various FSL methods, with either human-annotated or auto-annotated attributes. 
    Experimental results reveal a \emph{}{linear relationship} between few-shot performance and TAD distance.
    \item We demonstrate that our TAD metric can be applied to solve real-world FSL problems effectively: data augmentation and test-time intervention.
\end{itemize}


\section{Related works}
\subsection{Few-Shot Learning}
Few-Shot Learning is a challenging machine learning problem, where the key is to overcome the unreliable empirical data distribution caused by few samples. To address this challenging problem, several methods are proposed, spanning from data augmentation and meta-learning to transfer-learning. 

\textbf{Data Augmentation. }
To tackle the FSL problem, a straightforward approach involves augmenting the available data. 
When there are sufficient data for novel tasks, FSL is expected to return to the conventional machine learning paradigm. 
A widely used method is to transform various rules on the data \cite{lecun1998gradient}, involving horizontal and vertical translation, squeezing, scaling, cropping, Gaussian blurring, and so on. 
However, the efficacy of the traditional data augmentation method is limited when the support data quality is subpar.
Thus, some works attempt to generate more training data by overlapping semantic information from different images \cite{yun2019cutmix, chen2019IDeMe-Net}, inpainting the erased images \cite{li2021erasing-inpainting} and transforming inter-class or intra-class variations \cite{miller2000congealing, hariharan2017SGM, antoniou2017DAGAN, schwartz2018delta-encoder}.
Other works \cite{li2020AFHN, zhang2021PC, xu2022rsvae, an2023IDEAL} focus on the feature level, and aim to learn a series of feature augmentations to enrich the feature diversity and discriminative information by transforming the inter-class or intra-class variations on features.
Except for the above methods, recent works \cite{yang2020dc, yang2021dc2, wei2023ddwm} focus on calibrating the inaccurate feature distribution through related statistical information.
It is assumed that the training and novel categories are similar enough thus their means and variances could be shared to a large extent.
Building on the assumption, these methods calibrate the means and variances of novel categories by utilizing the distribution information of training categories.
In this paper, we present a potential application of our proposed TAD metric: augmenting the data of novel test tasks by translating the statistics from the most related training tasks.

\textbf{Meta-Learning. }
A more popular approach is meta-learning, which aims to learn cross-task knowledge by customizing the learning paradigm.
These meta-learning based methods \cite{vinyals2016matchnet, finn2017maml, li2017meta-sgd, snell2017prototypical, sung2018relationnet, hou2019CAN, allen2019IMP, sun2019mtl,  antoniou2019trainmaml, chen2019baseline++, zhang2020deepemd, fei2021z-norm, chen2021meta-baseline, afrasiyabi2022SetFeat, xie2022DeepBDC, zhang2022deepemd2} can be roughly divided into two categories.
The metric-based approaches \cite{ vinyals2016matchnet, snell2017prototypical, sung2018relationnet, hou2019CAN, allen2019IMP, fei2021z-norm, afrasiyabi2022SetFeat, xie2022DeepBDC, hao2023cpea, ye2023leadnet} aim to learn a cross-task embedding function and predict the query labels based on the learned distances.
The optimization-based approaches focus on learning some optimization state, like model initialization \cite{finn2017maml, sun2019mtl, sun2020mtl-hard, wang2022metantk}, step sizes \cite{li2017meta-sgd, antoniou2019trainmaml} or loss functions \cite{baik2021metal}, to rapidly update models with very few labeled samples.
In parallel to the empirical success of meta-learning methods, a series of theoretical works study how meta-learning utilizes the knowledge obtained from the training tasks and generalizes to the novel test tasks.
Many works \cite{pentina2014pac, amit2018meta-pac, rothfuss2021pacoh, ding2021bridging, guan2022fast} give a generalization error bound on novel tasks from the PAC-Bayesian perspective \cite{mcallester1998pac-bayes, mcallester1999pac}. 
These works often assume that each task is sampled from a meta task distribution.
Under such assumption, the generalization error bound on novel tasks can be reduced with increasing number of training tasks and training samples \cite{amit2018meta-pac}.
Some recent works replace the meta task distribution assumption with other conditions.
For example, \cite{cao2019shots, tripuraneni2021provable, du2020representation} assume a common representation function between different tasks, 
based on which the sample complexity bound for the novel task is derived.
Besides the strong assumptions, the above works do not quantify the relationship between training and novel tasks, and seldom explore the adaptation difficulty on novel tasks.
In this paper, we propose the TAD as a metric to quantify the task relatedness, and provide a new generalization error bound on a novel task.

\textbf{Transfer-Learning. }
Many recent works \cite{chen2019baseline++, chen2021meta-baseline, luo2021cosoc, hu2022pmf, dong2022sun_method, hiller2022fewture, lin2023smkd} have shown that standard transfer-learning procedure of early pre-training and subsequent fine-tuning is a strong baseline for few-shot learning. 
Without the complicated design of meta-learning strategies, a simple baseline method with deep backbone networks can achieve comparable or better performance than the state-of-the-art meta-learning methods \cite{chen2019baseline++, luo2021cosoc, triantafillou2020meta-dataset, guo2020broader}.
Recent transfer-based methods \cite{caron2021dino, dong2022sun_method, hiller2022fewture, lin2023smkd} focus more on learning embedding with good generalization ability.
FewTURE \cite{hiller2022fewture} splits the input samples into patches and encodes these patches through the Vision Transformers.
The Vision Transformers is first pre-trained with self-supervised iBOT \cite{zhou2021ibot}, and then uses inner loop patch importance re-weighting for supervised fine-tuning.
SMKD \cite{lin2023smkd} focuses on the gap between objectives of self-supervised learning and supervised learning, and then designs two supervised-contrastive losses on both class and patch levels to fill the gap. 
Beyond the aforementioned works employing self-supervised learning methods, an alternative approach involves leveraging information from textual modalities to acquire good embedding. 
In the first step, the multimodal models \cite{radford2021CLIP, tsimpoukelli2021frozen, li2022blip, li2023blipv2, najdenkoska2023meta-mapper}, jointly train the text backbone and visual backbone with multimodal data.
Subsequently, prefix tuning \cite{tsimpoukelli2021frozen} or prompt tuning \cite{zhou2022coop, zhou2022cocoop, najdenkoska2023meta-mapper} is employed to adapt the models to novel tasks.
In this work, we apply the proposed TAD metric into some transfer-based methods to measure the adaptation difficulty on novel tasks, and demonstrate the effectiveness of TAD metric.

\subsection{Task Relatedness and Difficulty Measure}
Quantifying the task relatedness and adaptation difficulty is fundamental and important for understanding the transferable knowledge in FSL.
Based on that, we can establish a principled approach for reusing the knowledge among related tasks, identifying tasks that exhibit effective transfer to any given target task, or intervening in these challenging target tasks with minimal cost.

\textbf{Task Relatedness. }
Several studies \cite{guo2020broader, triantafillou2020meta-dataset, song2023fsl_survey23} have demonstrated that numerous FSL methods exhibit effectiveness only when the training and novel test tasks are closely related.
However, the question that how to quantify the task relatedness has not been explored sufficiently in the FSL field.
One class of approaches measure the category relationship within a common representation space.
With properly learned representation function, various distribution divergence metrics, such as EMD \cite{rubner1998emd}, FID \cite{heusel201fid}, can be employed to calculate the distance between feature distributions of different categories, as well as tasks over them.
The primary challenge in this approach lies in learning a good representation function for novel categories with only a few labeled samples.
Previous works \cite{milbich2021ooDML, oh2022domain-similarity, liu2022wasserstein} assume that the representation function learned from training categories can be directly applied to novel categories, but this assumption may not hold due to significant category gaps.
Inspired by previous transfer learning works \cite{zamir2018taskonomy, achille2019task2vec}, recent study \cite{le2022task-affinity}  utilizes the Fisher Information Matrix to quantify the task relatedness without assuming a common representation function.
However, this approach has a heavy burden of computing Hessian matrix.
Besides, the calculated task distances are highly dependent on the learned model, making them difficult to apply to other FSL models.
In this paper, we introduce TAD to quantify the task relatedness and overcome the existing obstacles via attributes.
Unlike the above methods, TAD only relies on the attribute distributions, making it independent of models and easy to compute.

\textbf{Task Difficulty. }
The task difficulty has been explored in FSL from two aspects: measure the difficulty of (1) training tasks and (2) novel tasks. For training tasks, many previous works \cite{sun2019mtl, liu2020ada-sample, sun2020mtl-hard, arnold2021uniform-sample} have attempted to measure their difficulties based on a model’s output, such as negative log-likelihood or accuracy, and use this information to sample different tasks to train the model. The sampling strategy based on training task difficulty is similar to the area of hard example mining \cite{shrivastava2016hard-sample-mining} or curriculum learning \cite{bengio2009curriculum} that trains a model according to a specific order of samples to improve its generalization ability. For novel tasks, recent works \cite{achille2019task2vec, ren2020fsal, milbich2021ooDML, sariyildiz2021concept-gen, le2022task-affinity, oh2022domain-similarity, jiang2022fsir} try to measure the adaptation difficulty of novel tasks based on the relationship between novel and training tasks. 
These works utilize specific metrics, such as EMD and FID, to quantify the task relatedness, then select the most related training data to fine-tune models \cite{le2022task-affinity} or construct more challenging benchmarks to explore the generalization of models \cite{ren2020fsal, sariyildiz2021concept-gen, oh2022domain-similarity, jiang2022fsir}. Closer to our work, \cite{ren2020fsal} introduces a score to quantify the transferability of novel tasks via attributes and explore the generalization of FSL models at different transferability scores.
However, different from our work, \cite{ren2020fsal} aims to investigate the benefits of self-supervised pre-training with supervised fine-tuning in the few-shot attribute classification context.

\section{Task Attribute Distance}

The key to answering the first question we raised above lies in a proper metric that quantifies the relatedness between training and novel tasks. 
In this section, we begin by describing the FSL problem setting and then introduce the Task Attribute Distance (TAD) to quantify task relatedness via attributes.
The TAD metric first measures the category relationship through attribute conditional distributions, based on which the relationship between tasks is measured as well.
Finally, we discuss practical considerations in computing the TAD on real data.

\subsection{Problem Setting}
In few-shot learning problem, a model observes $n$ different training tasks $\{\tau_i\}_{i=1}^n$ with each task represented as a pair $\tau_i=(\mathcal{D}_i, S_i), 1\leq i\leq n$. 
$\mathcal{D}_i$ is an unknown data distribution over the input space $\mathcal{X}$ and label space $\mathcal{Y}_i$. $S_i=\{(x_k, y_k)|(x_k,y_k)\sim \mathcal{D}_i\}_{k=1}^{m_i}$ represents an observed training set drawn from $\mathcal{D}_i$.  
With a model trained on the $n$ training tasks, our target is to adapt and evaluate it on $t$ novel test tasks $\tau_{j}'=(\mathcal{D}_{j}',S_{j}')$, $1\leq j\leq t$. 
$\mathcal{D}_{j}'$ is an unknown data distribution over $\mathcal{X}$ and $\mathcal{Y}_{j}'$, and $S_{j}'$ is a labeled dataset drawn from $\mathcal{D}_j'$.  
Note that, in few-shot learning, the labeled data for each category is very limited, and the categories in training tasks $\{\tau_i\}_{i=1}^n$ will not appear in novel tasks $\{\tau_j'\}_{j=1}^t$, which means $ \mathcal{Y}_i\cap\mathcal{Y}'_j = \emptyset$,  $\forall i\in\{1, ..., n\},j\in\{1, ..., t\}$.

\subsection{Measuring Task Relatedness via Attributes}
To measure the relationship between two categories $y_k$ and $y_t$, a natural idea is to measure the divergence of class-conditional distributions.
An intuitive divergence measure is the $L_1$ or total variation distance \cite{gibbs2002tv_dis}, which is defined as
\begin{align}
    d(y_k, y_t) &= d_{TV}(p(x|y_k), p(x|y_t)) \notag\\
    &= \sup_{B\in\mathcal{B}}\left| P_{x|y_k}[B]-P_{x|y_t}[B]\right|,
\end{align}
where $\mathcal{B}$ is the set of measurable subsets under distributions $p(x|y_k)$ and $p(x|y_t)$, and $P_{x|y_k}[B]$ denotes the probability of subset $B\in\mathcal{B}$ under $p(x|y_k)$. 
Unfortunately this distance is often hard to compute, because the distribution of $p(x|y)$ is usually unknown and the set $\mathcal{B}$ is often infinite.

In this paper, we introduce attributes to overcome the above limitation because the values of attributes are usually discrete and finite.
The basic idea is based on the \emph{attribute compositionality hypothesis} presented in the introduction section. 
Specifically, the distance between two categories can be measured by the difference between the corresponding attribute conditional distributions.
Follow the above idea, let $\mathcal{A}$ be the attribute space spanned by $L$ attribute variables $\{a^l\}_{l=1}^L$, we define the \emph{distance between two categories} $y_k$ and $y_t$ on $L$ attribute variables as:
\begin{align}
    d_{\mathcal{A}}(y_k, y_t) &= \frac{1}{L}\sum_{l=1}^L d_{TV}(p(a^l|y_k), p(a^l|y_t)) \\ 
     &= \frac{1}{L} \sum_{l=1}^L \sup_{B\in\mathcal{B}^l}\left| P_{a^l|y_k}[B]-P_{a^l|y_t}[B]\right|,
    \label{eq:dis_category}
\end{align}
where $\mathcal{B}^l$ is the set of measurable subsets under attribute conditional distributions $p(a^l|y_k)$ and $p(a^l|y_t)$, and $\mathcal{B}^l$ is a finite set if the values of variable $a^l$ are finite.

Based on the distance between categories, we then define the distance metric between training and novel test tasks.
To this end, we represent the set of categories in training task $\tau_i$ and the set of categories in novel task $\tau_{j}'$ as two disjoint vertex sets respectively, and construct a weighted bipartite graph $G$ between the two vertex sets, where each node denotes a category and each edge weight represents the distance between two categories in $\tau_i, \tau_{j}'$ respectively.
Let $M=\{e_{kt}\}$ denote a maximum matching of  $G$, which contains the largest possible number of edges and any two edges do not connect the same vertex.
We choose $M$ with minimum weights and define the \emph{task distance} as
\begin{align}
    d(\tau_i, \tau_{j}') &= \frac{1}{\left|M\right|} \sum_{e_{kt}\in M} d_{\mathcal{A}}(y_k, y_t) \\
    &=  \frac{1}{L\left|M\right|} \sum_{e_{kt}\in M} \sum_{l=1}^L d_{TV}(p(a^l|y_k), p(a^l|y_t)),
    \label{eq:dis_task}
\end{align}
where $\left|M\right|$ is the number of edges in matching $M$.
From this definition, if the two tasks $\tau_i, \tau_{j}'$ are identical, the task distance $d(\tau_i, \tau_{j}')$ is equal to 0.
We call the task distance $d(\tau_i, \tau_{j}')$ as the Task Attribute Distance (TAD).
Additionally, the TAD is invariant to the permutation of categories in tasks.
In other words, modifying the numeric order of categories in $\tau_i$ or $\tau_{j}'$ does not affect their distance $d(\tau_i, \tau_{j}')$.
To emphasize the attribute conditional distributions with respect to different tasks, we hereinafter add task indexes on the distribution and probability notations, like $p^i(a^l|y_k)$ and $P^i_{a^l|y_k}[B]$, although the true distributions are task agnostic.

\subsection{Distance Computation and Approximation}
Finally, we discuss practical considerations in computing the TAD on real data, where the attributes often take on discrete and finite values. 
For continuous and infinite attributes, it is possible to divide their values into many segments or discrete parts.
Denote $V^l$ as a finite set of possible values of attribute $a^l$, we can re-express the TAD in Eq. (\ref{eq:dis_task}) as:
\begin{align}
    d(\tau_i, \tau_{j}') 
    &= \frac{1}{2L\left|M\right|} \sum_{e_{kt}\in M} \sum_{l=1}^L \sum_{v\in V^l}\left| P^i_{a^l|y_k}[v]-P^j_{a^l|y_t}[v]\right|. 
    \label{eq:origin}
\end{align}

Computing the above distance is still challenging as it requires finding the maximum matching $M$ with minimum weights, which is a combinatorial optimization problem.
The Hungarian algorithm \cite{kuhn1955hungarian} is commonly used to solve the matching problem, but it is computationally expensive.
Due to the high computational cost, we do not calculate this distance directly but estimate the approximation instead: 
\begin{equation}
    d(\tau_i, \tau_{j}') \approx \frac{1}{2LC} \sum_{l=1}^L \sum_{v\in V^l}\left| \sum_{k=1}^C P^i_{a^l|y_k}[v]-\sum_{t=1}^C P^j_{a^l|y_t}[v]\right|, 
    \label{eq:approximation}
\end{equation}
where $C$ denote the number of categories in task $\tau_i$ and $\tau_{j}'$.
In the Eq. (\ref{eq:approximation}), we simplify the comparison of attribute conditional distributions between two tasks by replacing the individual differences with the average difference.
Through the above approximation, we can avoid seeking the minimum weight perfect matching $M$ and estimate the distance efficiently.

To illustrate the differences between the original TAD as defined in Eq. (\ref{eq:origin}) and approximate TAD as defined in Eq. (\ref{eq:approximation}), we compare their performance and computational complexity in our experiments.
In all experiments, unless explicitly specified, we use approximate distance, as we have found it to yield similar results but with more efficient computations. 
Note that the theoretical analysis in the next section does not involve specific computations.
Hence, for our theoretical derivation, we utilize the original TAD metric defined in Eq. (\ref{eq:dis_task}).


\section{Theoretical Analysis on Generalization}
We have defined TAD metric to quantify the relatedness between tasks, further question is how the task relatedness affects the adaptation difficulty on novel tasks for different FSL models?
We try to explore this question in this section through theoretical analysis on generalization.
We first introduce a meta-learning framework with attribute learning to facilitate our theoretical analysis.
Then, we provide theoretical proof that establishes a connection between the proposed TAD metric and the generalization error bound on a novel task.

\subsection{A Specific Few-Shot Learning Framework}
\label{Sec 4.2.1}
\textcolor{black}{
To facilitate the subsequent theoretical analysis, we consider a meta-learning framework with attribute learning.
In this framework, a model is composed of two parts: an embedding function $f_\theta: \mathcal{X}\rightarrow \mathcal{A}$ parameterized by $\theta$ learns the mapping from a sample $x\in\mathcal{X}$ to attributes $a\in\mathcal{A}$, and a prediction function $g_{\phi}: \mathcal{A}\rightarrow \mathcal{Y}$ parameterized by $\phi$ learns the mapping from attributes $a\in\mathcal{A}$ to a class label $y\in\mathcal{Y}$ for each task.
During training, $f_\theta$ and $g_\phi$ are meta-learned from $n$ different training tasks.
To adapt to a novel task $\tau_{j}'$, we train the task-specific base-learner $f_{\theta_{j}'}$ and $g_{\phi_{j}'}$ based on few labeled samples $S_j'$ and meta-learned parameters $\theta, \phi$. }

In the following theoretical deduction, we will denote $p_{\theta_i}(a,y)\triangleq p(f_{\theta_{i}}(x),y)$ as a joint distribution over $\mathcal{A}\times\mathcal{Y}_i$ induced by the task-specific mapping $f_{\theta_i}$ for task $\tau_i$. 
Further, with the induced conditional distribution $p_{\theta_i}(a|y)$, we can compute the distance between task $\tau_i$ and $\tau_j'$ as 
\begin{align}
    d_\theta(\tau_i, \tau_{j}') &= \frac{1}{L\left|M\right|}\sum_{e_{kt}\in M} \sum_{l=1}^L d_{TV}(p_{\theta_i}(a^l|y_k), p_{\theta_{j}'}(a^l|y_t)).
\end{align}
Note that $d(\tau_i, \tau_j')$ (as defined in Eq. (\ref{eq:dis_task})) computed based on $p(a|y)$ is the model-agnostic distance metric, while $d_\theta(\tau_i, \tau_j')$ is a model-related one since it relies on the learned mapping $f_{\theta_i}$ and $f_{\theta_{j}'}$ for training and novel task $\tau_i$, $\tau_{j}'$, respectively.

\subsection{Measuring Task Adaptation Difficulty via Attributes}
As for the above meta-learning framework, we provide theoretical analysis of the generalization error on a novel task, in terms of the proposed TAD metric.
We define the generalization error and empirical error of the meta-learned parameters $\theta$ and $\phi$ on novel task $\tau_{j}'$ as
\begin{align*}
    \epsilon(\theta, \phi; \tau_{j}')&=\mathbb{E}_{(x,y)\sim \mathcal{D}_{j}'}[\mathbb{I}(g_{\phi_{j}'}(f_{\theta_{j}'}(x))\neq y)], \\
    \hat{\epsilon}(\theta, \phi; \tau_{j}')&=\frac{1}{m_{j}'} \sum_{k=1}^{m_{j}'}\mathbb{I}(g_{\phi_{j}'}(f_{\theta_{j}'}(x_k))\neq y_k), \notag
\end{align*}
where $\theta_{j}', \phi_{j}'$ are the parameters of task-specific base-learner $f_{\theta_{j}'}, g_{\phi_{j}'}$, which are learned from labeled samples $S_{j}'$ given meta-learned parameters $\theta$ and $\phi$.
${m_j'}=\left|S_{j}'\right|$ is the number of labeled samples in $S_{j}'$, and $\mathbb{I}$ denotes the indicator function.
Similarly, we can define the generalization error $\epsilon(\theta, \phi; \tau_i)$ and empirical error $\hat{\epsilon}(\theta, \phi; \tau_i)$ for training task $\tau_i$.
With these definitions, we will proceed by introducing an auxiliary lemma, and then stating our main theoretical results.
All detailed proofs of the following lemma and theorems are in the Appendix A. 

\begin{lemma}
    Let $\mathcal{A}$ be the attribute space, $L$ be the number of attributes.
    Assume all attributes are independent of each other given the class label, i.e. $p(a|y) = \prod_{l=1}^Lp(a^l|y)$.
    For all $a_i \in \mathcal{A}$ and any two categories $y_k, y_t$, the following inequality holds:
    \begin{equation}
        \sum_{a_i\in\mathcal{A}}\left|p(a_i|y_k)-p(a_i|y_t)\right| \leq d_{\mathcal{A}}(y_k, y_t) +  \Delta,
        \label{eq:distance ineq}
    \end{equation}
    where $d_{\mathcal{A}}(y_k, y_t)$ is the distance as defined in Eq.(\ref{eq:dis_category}) and $\Delta=\sum_{a_i\in\mathcal{A}}\frac{1}{2L}\sum_{l=1}^L(p(a^l_i|y_k)+p(a^l_i|y_t))$.
    \label{lemma 1}
\end{lemma}
Lemma \ref{lemma 1} says that under the attribute conditional independence assumption, the distance between two attribute conditional distributions $p(a|y_k), p(a|y_t)$ over attribute space $\mathcal{A}$ is 
bounded by the defined distance $d_{\mathcal{A}}(y_k, y_t)$ and a non-negative term $\Delta$.
This result enables us to make use of the defined TAD metric in deriving the generalization error bound, leading to the following theoretical results.

\begin{theorem}
    With the same notation and assumptions as in Lemma \ref{lemma 1}, 
    let $\mathcal{H}$ be the hypothesis space with VC-dimension $d$, 
    $f_\theta$ and $g_\phi$ be the embedding function and prediction function as introduced in Sec. \ref{Sec 4.2.1} respectively. 
    Denote $g_{\phi^*}$ as the best prediction function on some specific tasks given a learned embedding function.
    For any single training task $\tau_i=(\mathcal{D}_i, S_i)$ and a novel task $\tau_{j}'=(\mathcal{D}_{j}', S_{j}')$, suppose the number of categories in the two tasks is the same, then with probability at least $1-\delta$, $\forall g_\phi\circ f_\theta\in\mathcal{H}$, we have 
    \begin{align}
         \epsilon(\theta, \phi; \tau_{j}')&\leq 
            \hat{\epsilon}(\theta, \phi; \tau_i) + \sqrt{\frac{4}{m_i}(d\log\frac{2em_i}{d}+\log\frac{4}{\delta})}
            \notag \\ &+ d_\theta(\tau_i, \tau_{j}') + \Delta' + \lambda, 
    \end{align}
    where $\lambda=\lambda_i+\lambda_j'$ is the generalization error of $g_{\phi^*}$ on the two tasks, i.e., 
    $\lambda_i=\mathbb{E}_{(x,y)\sim \mathcal{D}_i}[\mathbb{I}(g_{\phi_{i}^*}(f_{\theta_i}(x))\neq y)]$, $\lambda_j'=\mathbb{E}_{(x,y)\sim \mathcal{D}_{j}'}[\mathbb{I}(g_{\phi_{j}'^*}(f_{\theta_j'}(x))\neq y)]$. $\Delta'$ is a term depending on learned prediction functions $g_{\phi_i}, g_{\phi_j'}$ and the best prediction functions $g_{\phi_i^*}, g_{\phi_j'^*}$.
    \label{theorem 1}
\end{theorem}

Theoretically, the generalization error on a novel task is bounded by the the training task empirical error $\hat{\epsilon}(\theta, \phi; \tau_i)$ plus four terms: 
the second term $d_\theta(\tau_i, \tau_{j}')$ is the model-related distance between $\tau_i$ and $\tau_{j}'$, which is derived based on Lemma \ref{lemma 1};
the third term $\Delta'$ reflects the classification ability of $g_\phi$, which converges to zero if the learned prediction functions are equal to the best ones for both tasks;
the last term $\lambda$ is the generalization error of $f_\theta$ and $g_{\phi^*}$, which depends on the attribute discrimination ability of $f_\theta$ and the hypothesis space of prediction function $g_\phi$.
For a reasonable hypothesis space, if $f_\theta$ has a good attribute discrimination ability on both tasks, the last term usually converges to zero. 
Next we generalize this bound to the setting of $n$ training tasks.

\begin{corollary}
   With the same notation and assumptions as Theorem 1, 
    for $n$ training tasks $\{\tau_i\}_{i=1}^n$ and a novel task $\tau_{j}'$, 
    define $\hat{\epsilon}(\theta, \phi; \tau_{i=1}^n)=\frac{1}{n}\sum_{i=1}^n \hat{\epsilon}(\theta, \phi; \tau_i)$, then with probability at least $1-\delta$, $\forall g_\phi\circ f_\theta\in\mathcal{H}$, we have
    \begin{align}
        \epsilon(\theta, \phi; \tau_{j}')&\leq 
            \hat{\epsilon}(\theta, \phi; \tau_{i=1}^n) + \frac{1}{n}\sum_{i=1}^n\sqrt{\frac{4}{m_i}(d\log\frac{2em_i}{d}+\log\frac{4}{\delta})}
            \notag \\ 
            & + \frac{1}{n}\sum_{i=1}^n d_\theta(\tau_i, \tau_{j}') + \Delta' + \lambda, 
            \label{eq:corollary 1}
    \end{align}
    where $\lambda=\frac{1}{n}\sum_{i=1}^n \lambda_i + \lambda_j'$, and $\Delta'$ is a term depending on the learned prediction functions $\{g_{\phi_i}\}_{i=1}^n, g_{\phi_j'
    }$ and the best prediction functions $\{g_{\phi_i^*}\}_{i=1}^n, g_{\phi_j'^*}$.
    \label{corollary 1}
\end{corollary}
Corollary \ref{corollary 1} is a straightforward extension of Theorem \ref{theorem 1}, in which we consider multiple training tasks instead of a single training task.
In Corollary \ref{corollary 1}, the generalization error on a novel task $\tau_{j}'$ is  bounded partially by the average distance between task $\tau_{j}'$ and $n$  training tasks.
Note that the distance $d_\theta(\tau_i, \tau_j')$ used in the bound is 
model-related.
Next, we further derive the relationship between model-related distance and model-agnostic distance as follows.

\begin{definition}[\textcolor{black}{$\xi$-approximation meta-mapping}]
    \label{definition 1}
    \textcolor{black}{An embedding function $f_{\theta}:\mathcal{X}\rightarrow\mathcal{A}$, parameterized by $\theta$, is termed an $\xi$-approximation meta-mapping for $n$ training tasks $\{\tau_i\}_{i=1}^n$ if $\|f_{\theta_i}-f_{\theta_i^*}\|<\xi$,
    where $f_{\theta_i}$ is the task-specific embedding function for task $\tau_i$ based on training set $S_i$ and meta parameter $\theta$, and $f_{\theta_i^*}$ is the best task-specific embedding function.}
\end{definition}

\begin{theorem}
   \label{theorem 2}
    With the same notation and assumptions as in Corollary \ref{corollary 1}, assume that the conditional distribution $p(x|a^l)$ is task agnostic, 
    \textcolor{black}{and the embedding function $f_{\theta}$ is a $\xi$-approximation meta-mapping for $n$ training tasks.
    If $\xi$ tends to zero, the following equality holds:}
    \begin{align}
            \frac{1}{n}\sum_{i=1}^nd_\theta(\tau_i, \tau_{j}')&\leq\frac{1}{n}\sum_{i=1}^nd(\tau_i, \tau_{j}').
    \end{align}
\end{theorem}

\textcolor{black}{Theorem \ref{theorem 2} shows that when the task-specific embedding function $f_{\theta_i}$ is in the proximity of the best embedding function $f_{\theta_i^*}$ for $n$ training tasks, the model-related average distance $\frac{1}{n}\sum_{i=1}^nd_\theta(\tau_i, \tau_{j}')$ can be bounded by the average TAD $\frac{1}{n}\sum_{i=1}^nd(\tau_i, \tau_{j}')$.
With the same assumption as in Theorem \ref{theorem 2}, the third terms on the r.h.s. of Eq.~(\ref{eq:corollary 1}) in Corollary \ref{corollary 1} can be bounded by the average TAD, $\frac{1}{n}\sum_{i=1}^nd(\tau_i, \tau_j')$.
As the average TAD is model agnostic, we can rely on the TAD to measure the adaptation difficulty of each novel task, without the need for training any models.}

\textcolor{black}{Note that, many existing FSL methods can be regarded as specific instances of the meta-learning framework that we discuss above.
For instance, transfer-based methods involve keeping the embedding function $f_\theta$ fixed and only fine-tuning the prediction function $g_\phi$ during adaptation to novel tasks. 
Furthermore, the metric-based methods involve constructing the prediction function $g_\phi$ as a non-parametric metric function. 
The optimization-based methods, on the other hand, treat meta-parameters $\theta$ and $\phi$ as the initialization parameters for task-specific base-learners. 
However, the most significant difference in our considered meta-learning framework lies in the embedding function $f_\theta$, which maps images to a specific representation space, namely attribute space. 
Introducing attributes enable us to analyze how differently FSL models perform in adaptation to novel tasks, even if some of them do not learn attributes explicitly.}

\section{Experiments}
\label{sec:experiments}
In this section, we conduct experiments to validate our theoretical results and explore the generality of the proposed TAD metric in quantifying task relatedness and measuring task adaptation difficulty. 
Firstly, we evaluate our TAD metric in a scenario where human-annotated attributes are available and the training/novel tasks are sampled from the same dataset but with different categories. 
Next, we test the TAD in a more general and challenging scenario where human-annotated attributes are not available and the training tasks are constructed by sampling categories from  different datasets.

\subsection{Setups}
\textbf{Datasets:} 
We choose three widely used benchmarks:
(1) CUB-200-2011 (\textbf{CUB}) \cite{wah2011CUB}: CUB is a fine-grained dataset of birds, which has 200 bird classes and 11,788 images in total.
We follow \cite{kang2021renet} to split the dataset into 100 training classes, 50 validation classes and 50 test classes.
As a fine-grained dataset, CUB provides part-based annotations, such as beak, wing and tail of a bird.
Each part is annotated by a bounding box and some attribute labels.
Because the provided attribute labels are noisy, we denoise them by majority voting, as in \cite{koh2020conceptbottleneck}.
After the pre-processing, we acquire 109 binary category-level attribute labels.
(2) SUN with Attribute (\textbf{SUN}) \cite{patterson2014sun}: 
SUN is a scene classification dataset, which contains 14,340 images for 717 scene classes with 102 scene attributes.
In following experiments, We split the dataset into 430/215/72 classes for training/validation/test, respectively.
(3) \textbf{\emph{mini}ImageNet} \cite{vinyals2016matchnet}: \emph{mini}ImageNet is a subset of ImageNet consisting of 60,000 images uniformly distributed over 100 object classes. 
Note that, different with CUB and SUN, \emph{mini}ImageNet does not provide the attribute annotations, which means category-level attribute annotations are not available.
Following \cite{chen2019baseline++}, we consider the \emph{cross-dataset scenario} from \emph{mini}ImageNet to CUB, where we use 100 classes of \emph{mini}ImageNet as training classes, and the 50 validation and 50 test classes from CUB.

\begin{figure*}[!t]
\centering
\subfloat[]{\includegraphics[width=0.24\textwidth]{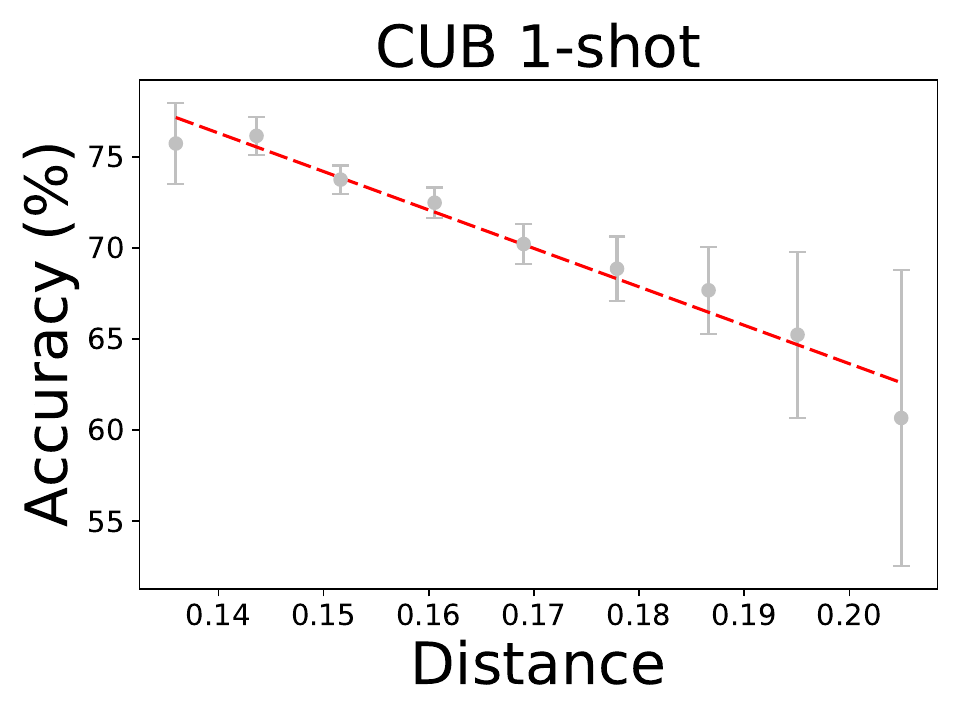}
\label{fig:apnet-a}}
\hfil
\subfloat[]{\includegraphics[width=0.24\textwidth]{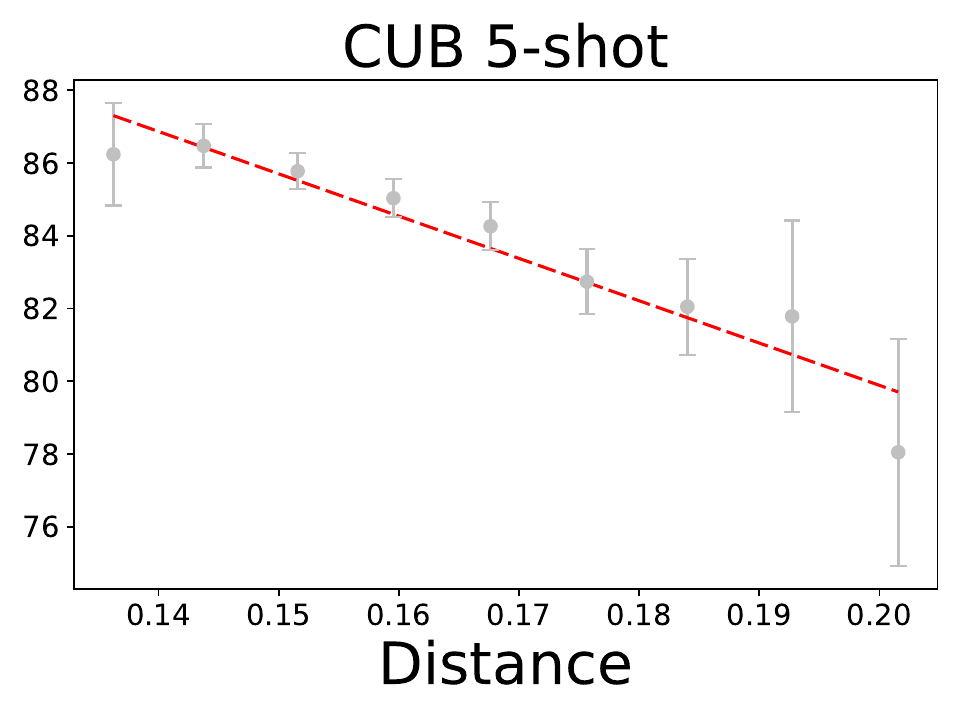}
\label{fig:apnet-b}}
\hfil
\subfloat[]{\includegraphics[width=0.24\textwidth]{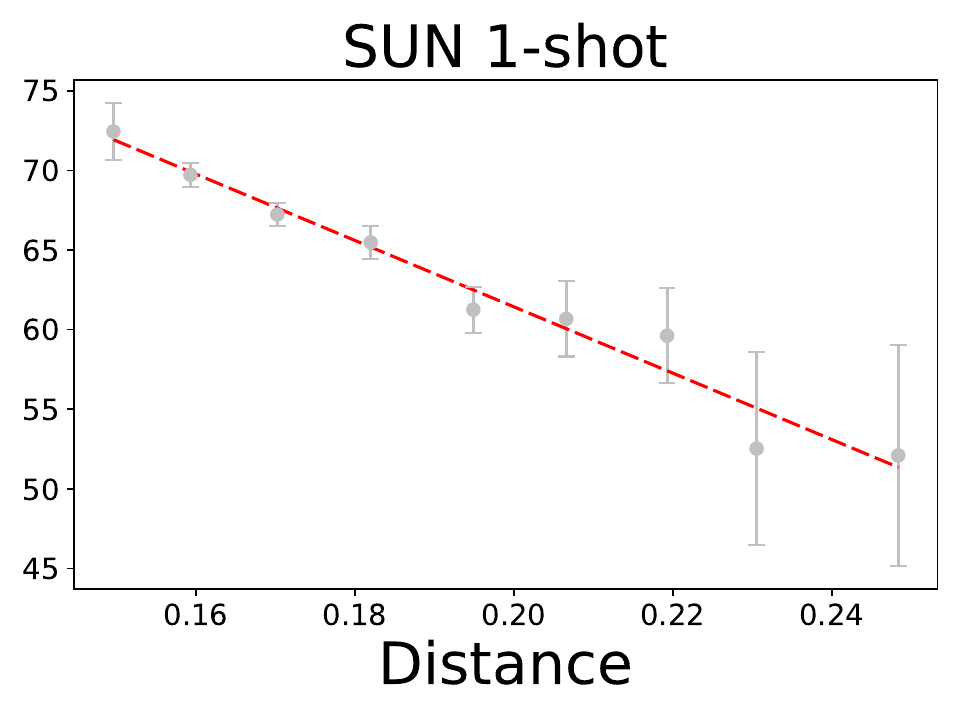}
\label{fig:apnet-c}}
\hfil
\subfloat[]{\includegraphics[width=0.24\textwidth]{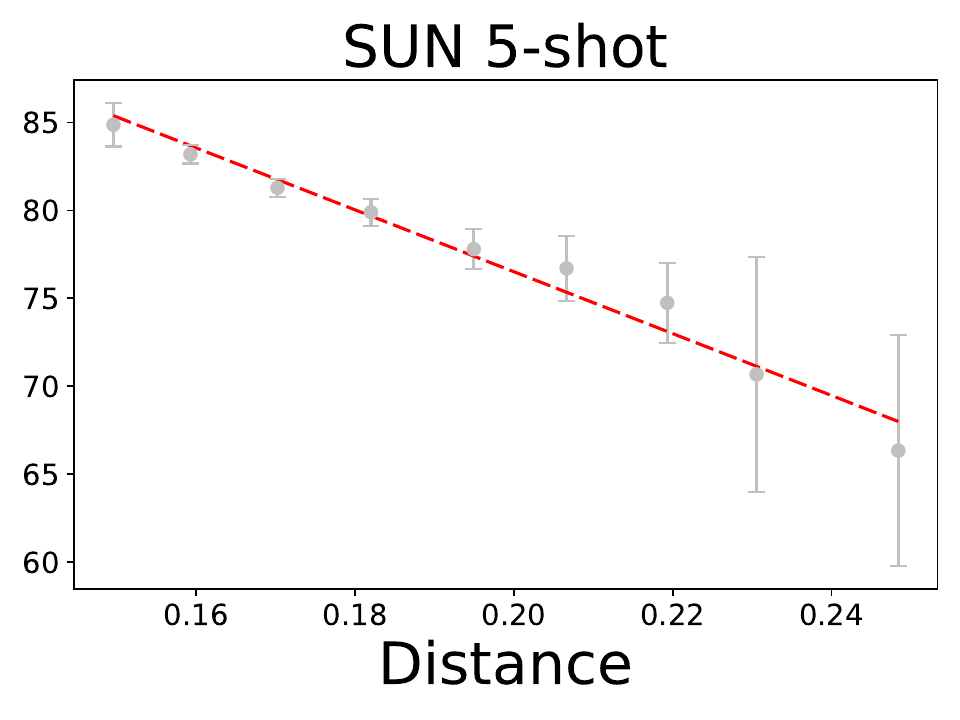}
\label{fig:apnet-d}}
\caption{Accuracy of APNet in terms of the average task distance. (a)-(b) 5-way 1-shot and 5-shot on CUB dataset. (c)-(d) 5-way 1-shot and 5-shot on SUN dataset. In (a)-(d), each gray point denotes the average accuracy for all points in a distance interval, and the error bar denotes the confidence interval at 95\% confidence level. The red dashed line is a fitted line, which shows the tendency of these gray points.}
\label{fig:distance and task accuracy}
\end{figure*}

\textbf{Attribute Prototypical Network:} 
Our theoretical analysis is based on a specific meta-learning framework with attribute learning (proposed in Sec \ref{Sec 4.2.1}).
Thus, we first instantiate a simple model under that framework as an example to verify our theory.
Specifically, we adopt a four-layer convolution network (Conv-4) with an additional MLP as the embedding function $f_\theta$. 
The convolutional network extracts feature representations from images, then the MLP takes features as input and predicts attribute labels. 
To adapt to a novel task, we will keep $f_\theta$ fixed, which means that meta-learned parameter $\theta$ is equal to the task-specific parameter $\theta_{j'}$.
For prediction function $g_{\phi_i}$ parameterized by $\phi_i$, we simply choose an non-parametric metric function like ProtoNet \cite{snell2017prototypical}, which takes the attributes outputted by $f_\theta$ as input to compute the cosine distance between test samples and attribute prototypes, then predicts the target label.
We call this method as \emph{Attribute Prototypical Network} (APNet).
We train APNet by simultaneously minimizing the attribute classification loss and the few-shot classification loss.
See details of APNet in the Appendix B.

\textbf{Other FSL Methods:}
Besides of APNet, we choose five classical FSL methods in the following experiments, since they cover a diverse set of approaches to few-shot learning: (1) Matching Network (MatchingNet) \cite{vinyals2016matchnet}, (2) Prototypical Network (ProtoNet) \cite{snell2017prototypical}, (3) Relation Network (RelationNet) \cite{sung2018relationnet}, (4) Model Agnostic Meta-Learning (MAML) \cite{finn2017maml} and (5) Baseline++ \cite{chen2019baseline++}.
Note that these FSL methods, unlike the above APNet, do not use attribute annotations during training.
See more implementation and experimental details in the Appendix C.

\subsection{Task Attribute Distance with Human-annotated Attributes}
\textbf{Linear relationship under limited samples:}
According to Theorem \ref{theorem 2}, if the embedding function $f_\theta$ is a $\xi$-approximation meta mapping and $\xi$ tends to zero for $n$ training tasks, the TAD serves as a metric to measure the adaptation difficulty on a novel task.
This approximation condition is difficult to maintain when learning parameter $\theta$, because the best embedding function $f_{\theta^*}$ is often unknown and the number of training samples is limited in real-world applications.
Here, we try to verify our theoretical results empirically.
We train our APNet on CUB and SUN dataset, then use the provided attribute annotations to calculate the average distance $\frac{1}{n}\sum_{i=1}^nd(\tau_i, \tau_j')$ between each novel task $\tau_{j}'$ and $n$ training tasks $\{\tau_{i}\}_{i=1}^n$, respectively. 
For simplicity, we sample $n=100,000$ training tasks to estimate the distances to $2,400$ novel tasks.
Following $N$-way $K$-shot setting, each task only contains $N*K$ labeled samples for training or fine-tuning models.
Fig. \ref{fig:distance and task accuracy} shows the task distance and the corresponding accuracy on novel tasks for APNet.
We can observe that as the distance increases, the accuracy of APNet decreases in both 5-way 1-shot and 5-way 5-shot settings on CUB and SUN.
These results verify that TAD can characterize model's generalization error on each novel task and measure the task adaptation difficulty effectively, even when the best embedding function is unknown and only limited labeled samples are available for training and fine-tuning models.
Interestingly, we also find a \textbf{linear relationship} between task distance and accuracy in Fig.  \ref{fig:distance and task accuracy}: as the distance increases, the accuracy of APNet decreases \textbf{linearly} in all datasets and settings.
Note that the confidence interval in Fig. \ref{fig:distance and task accuracy} is much larger for the last a few points.
We argue this is because these distance intervals contain fewer novel tasks, thus the average accuracy of novel tasks in the interval is more easily affected by random factors.

\textbf{The number of labeled samples:}
Additionally, in Fig. \ref{fig:distance and task accuracy}, when comparing 1-shot and 5-shot results with the same distance interval, we find that the increase of accuracy varies at different distance intervals.
For instance, on the CUB dataset (comparing Fig. \ref{fig:apnet-a} with Fig. \ref{fig:apnet-b}), when the task distance is 0.14, APNet shows an improvement of approximately 10\% in accuracy for the 5-shot setting over the 1-shot setting, whereas it shows an improvement of around 15\% for the distance of 0.20. 
This suggests that increasing the number of labeled samples is more effective for harder tasks.
One possible explanation is that as the task distance increases, less knowledge can be transferred from training tasks, making it harder for models to adapt to the novel task. 
Hence, more information from the novel task is required for adaptation, and the model’s performance get more benefit from extra samples.

\begin{figure*}[!t]
  \centering
  \includegraphics[width=1\textwidth]{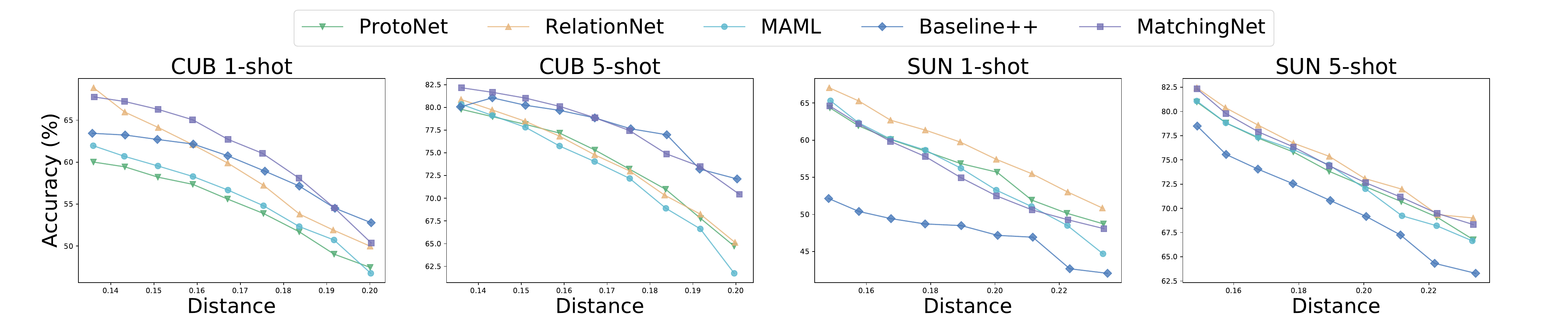}
  \caption{Accuracy of different methods in terms of the average task distance. From left to right, 5-way 1-shot and 5-shot on CUB/SUN.  Each point denotes the average accuracy in a distance interval.}
  \label{fig:distance and task accuracy of other methods}
\end{figure*}

\begin{figure*}[!t]
\centering
\subfloat[]{\includegraphics[width=0.24\textwidth]{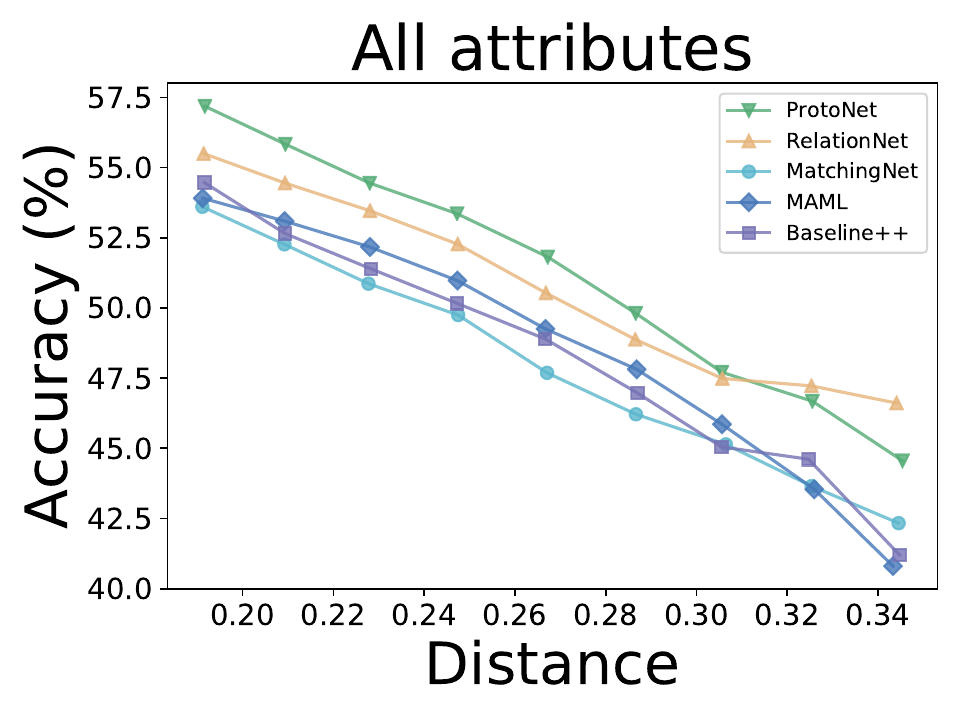}
\label{fig:cross-a}}
\hfil
\subfloat[]{\includegraphics[width=0.24\textwidth]{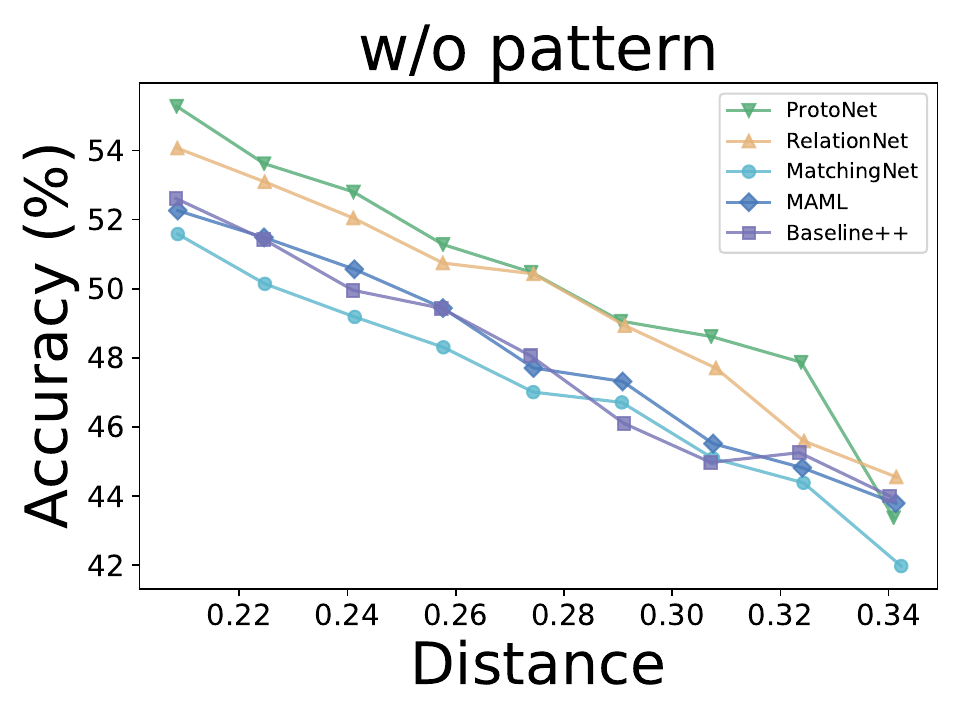}
\label{fig:cross-b}}
\hfil
\subfloat[]{\includegraphics[width=0.24\textwidth]{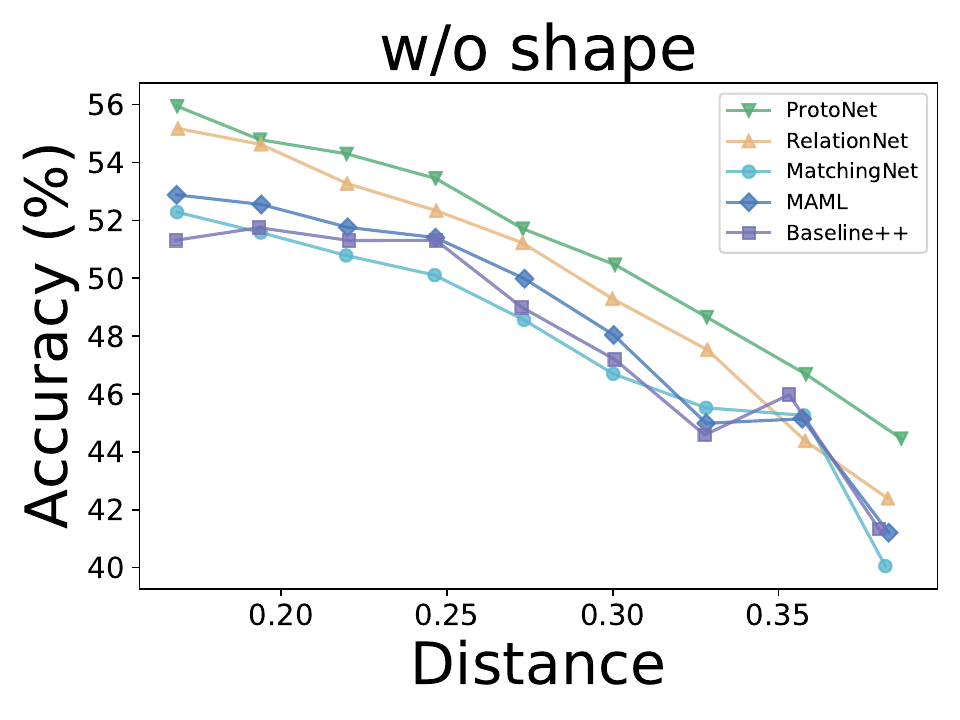}
\label{fig:cross-c}}
\hfil
\subfloat[]{\includegraphics[width=0.24\textwidth]{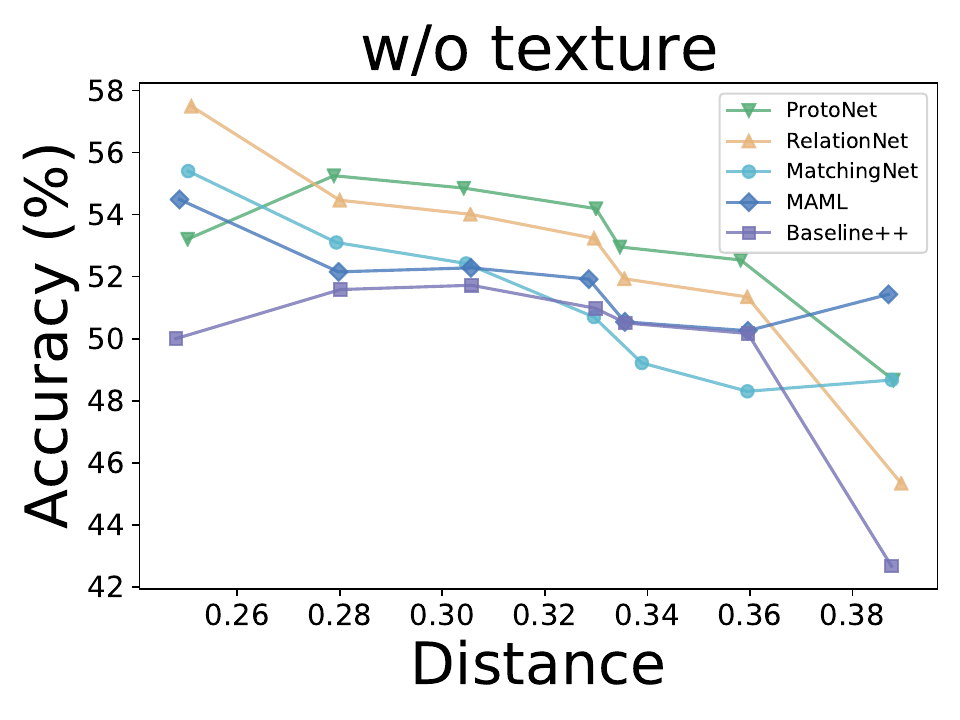}
\label{fig:cross-d}}
\caption{Accuracy of different methods in terms of the average task distance with (a) all attributes, (b) removal of pattern attributes, (c) removal of shape attributes, (d) removal of texture attributes. The experiment is conduct in 5-way 5-shot setting.}
\label{fig:distance and task accuracy autoann}
\end{figure*}

\textbf{Other FSL models:}
We have shown that the average TAD can effectively reflect the task adaption difficulty for APNet.
A natural question is whether the calculated task distance can be directly applied to other FSL methods, even if they do not follow the specific meta-learning framework that utilizes the attributes during training.
We try to empirically explore this question. 
With the same distance estimation and experimental setting, we conduct experiments with five different FSL methods on CUB and SUN.
Fig. \ref{fig:distance and task accuracy of other methods} shows the task distance and the corresponding accuracy of 2,400 novel test tasks for them.
We observe similar results that with the increase of task distance, the accuracy of all FSL models tends to decrease linearly.
This indicates that even though these FSL methods do not use the attribute annotations during training, they 
implicitly learn mixture of attributes in their representations. 
Therefore, attribute-based distance can still reflect the task adaptation difficulty for them. 
These results demonstrate the generality of the proposed TAD metric, and provide some insight into the transferable knowledge that different FSL models have learned from training tasks. 

\subsection{Task Attribute Distance with Auto-annotated Attributes}
\textbf{Attribute Auto-Annotation:}
Another interesting question is how to utilize the proposed TAD metric in situations where human-annotated attributes are either unavailable or expensive to obtain. 
One classic example is the cross-domain few-shot learning problem, since it requires the annotation of distinct datasets with a common attribute set.
To address this challenge, we propose a solution to auto-annotate attributes using a pretrained CLIP \cite{radford2021CLIP} model, which has demonstrated impressive zero-shot classification ability.
We follow previous work \cite{russakovsky2010attribute} and pre-define 14 attribute labels (See the details in Tab. \ref{tab:attribute list}) based on pattern, shape, and texture.
Then we create two descriptions for each attribute, such as ``a photo has shape of round'' and ``a photo has no shape of round'', which are used as text inputs for CLIP.
We formulate the annotation problem as 14 attribute binary classification problems for each image using CLIP model.
After that, we gather the attribute predictions of all images within the same category, which provides rough category-level attribute information while greatly reduces the cost of attribute annotations.
With the above auto-annotation process, we can obtain the category-level attributes for \emph{mini}ImageNet \cite{vinyals2016matchnet} and CUB \cite{wah2011CUB} in just 5 minutes.

\begin{table}
\centering
\caption{Details of pre-defined 14 attribute labels for auto-annotation, including pattern, shape and texture.}
\begin{tabular}{ll}
\toprule
        & \textbf{Attributes}                                                                \\ \midrule
Pattern & spotted, striped                                                          \\
Shape   & long, round, rectangular, square                                          \\
Texture & furry, smooth, rough, shiny, metallic, vegetation, wooden, wet            \\ \bottomrule
\end{tabular}
\label{tab:attribute list}
\end{table}

\textbf{Cross-dataset generalization:}
Follow previous works \cite{chen2019baseline++}, we consider a cross-domain few-shot learning scenario from \emph{mini}ImageNet to CUB.
We use the auto-annotation method described above to annotate the attributes for both \emph{mini}ImageNet and CUB dataset. 
We then train five FSL models on the cross-dataset scenario, and estimate the average task distance with the auto-annotated attributes.
Fig. \ref{fig:cross-a} illustrates the distance and corresponding accuracy of 2,400 novel test tasks with the auto-annotated attributes. 
We can find that, with the increase of task distance, the accuracy of different models tends to decrease.
This phenomenon is consistent with our previous findings and shows the proposed TAD metric still works with auto-annotated attributes.
Besides, we selectively remove some attributes to explore the influence of them in the distance-accuracy curve. 
Fig. \ref{fig:cross-b}, \ref{fig:cross-c}, \ref{fig:cross-d} show that the exclusion of pattern, shape, and texture attributes exhibits varying degrees of influence on the decreasing tendency. 
Notably, we discover that texture attributes are more importance than others, as indicated by the more pronounced fluctuations in the curve.

\begin{figure}
    \centering
    \includegraphics[width=0.4\textwidth]{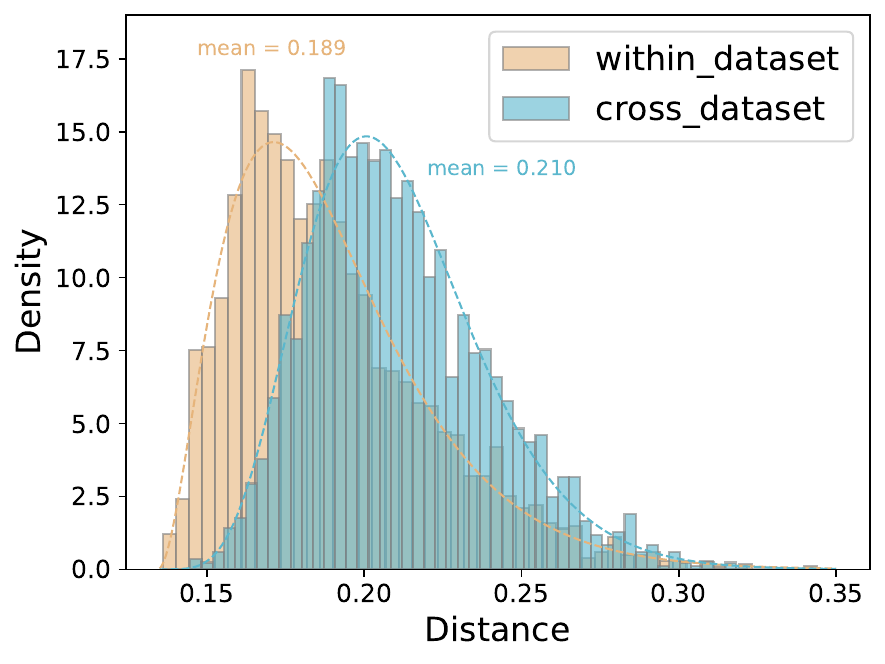}
    \caption{Distance distributions of novel tasks under two scenarios. Note that the distance distributions depend on the auto-annotated attributes and datasets, while are agnostic to specific FSL models.}
    \label{fig:distance distribution}
\end{figure}

\textbf{Cross-dataset distance:}
Next, we explore a common underlying hypothesis in many cross-domain few-shot learning (CD-FSL) works: the category gaps across dataset are usually larger than within a dataset, resulting in sampled training tasks less related to novel tasks.
Building on this hypothesis, the CD-FSL is considered as a more challenging problem.
We try to verify this hypothesis based on our TAD metric.
More specifically, we collect the distance distributions among 2,400 novel tasks under two scenarios: (1) within-dataset scenario: the training and novel tasks are sampled from the same dataset, although their categories are not overlapped; (2) cross-dataset scenario: the training and novel tasks are sampled from different datasets, and their categories are also not overlapped.
For within-dataset scenario, we split the total 100 classes of miniImageNet into 64/16/20 classes for training/validation/test, respectively.
For cross-dataset scenario, we use the same 64 classes of miniImageNet as training classes, and the 50 validation and 50 test classes from CUB.
We use the same auto-annotated attributes for TAD computation in the two scenarios.
Fig. \ref{fig:distance distribution} shows the distance distributions among 2,400 novel test tasks under the two scenarios.
We find that, the distance distributions under the two scenarios approximately follows the Gamma distributions.
However, the mean distance of cross-dataset scenario is much larger than that of within-dataset scenario.
The result demonstrates that although there are no class overlaps between training and novel tasks, tasks sampled from different dataset (e.g. miniImageNet and CUB) are less related than sampled from the same dataset (e.g. both from miniImageNet).
This may be one of the reasons that the same FSL models perform worse in the cross-dataset scenario.

\begin{table}[!t]
    \centering
    \begin{tabular}{lcc|c}
    \toprule
            & \textbf{100 classes} & \textbf{64 classes} & $\Delta$ \\ \midrule
    \textbf{MatchingNet} & 50.0        & \textbf{51.7}       & -1.7         \\
    \textbf{ProtoNet}    & \textbf{53.7}        & 52.6       & 1.1        \\
    \textbf{RelationNet} & \textbf{52.6}        & 50.3       & 2.3        \\
    \textbf{MAML}        & \textbf{51.6}         & 49.7      & 1.9         \\
    \textbf{Baseline++}  & 52.4        & \textbf{55.5}       & -3.1         \\ \bottomrule
    \end{tabular}
    \caption{Comparison of 5-shot accuracy for different methods, when training on selected 64 classes instead of all 100 classes. $\Delta$ represents the difference in accuracy between the two cases. The best accuracy of methods is marked in bold.}
    \label{tab:all vs part}
\end{table}

\textbf{Training with partial versus all tasks:}
Using the proposed TAD, we can also investigate whether training on all available training tasks leads to better generalization compared to training on only the most related ones. 
While it is generally expected that training with more data helps to improve generalization, it remains a question whether this holds for heterogeneous data. 
To explore this, we follow the experimental setup of the cross-dataset scenario and select these less related training tasks based on their calculated task distances. 
We collect the classes in the selected tasks, compute the frequency of each class, and remove the most frequent 36 classes (out of a total of 100 classes in \emph{mini}ImageNet). 
We then reconstruct training tasks based on the remaining 64 classes in \emph{mini}ImageNet and retrain five FSL models. The results are presented in Tab. \ref{tab:all vs part}. It can be observed that incorporating more heterogeneous data that is less related to the novel tasks may not lead to much improvement (
for ProtoNet, RelationNet and MAML), but could even result in performance degradation (
for MatchingNet and Baseline++). This result may inspire future research on how to better utilize heterogeneous data to improve the few-shot learning performance of models.

\begin{figure*}
    \centering
    \includegraphics[width=1.0\textwidth]{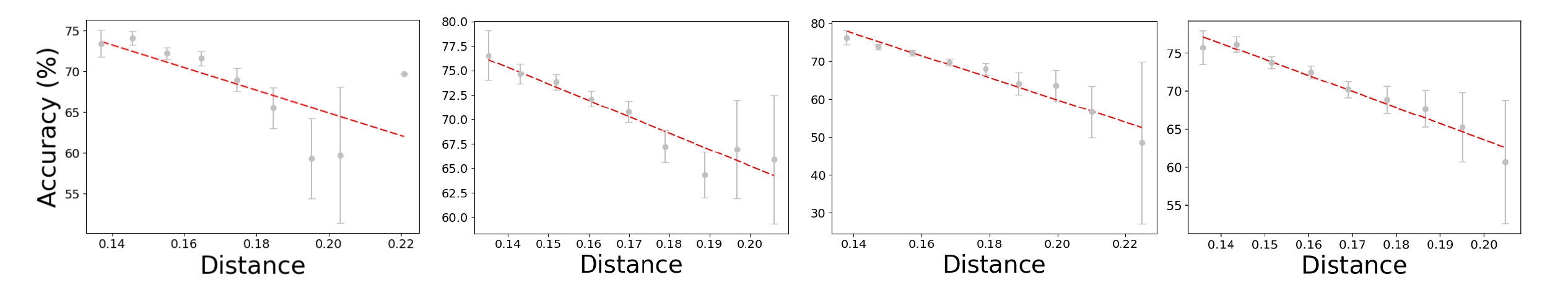}
    \caption{Accuracy of APNet in terms of the average task distance. From left to right, the value of $n$ is 100, 1000, 10000 or 100000. The experiments are conducted on CUB dataset under 5-way 1-shot setting.}
    \label{fig:sampled tasks}
\end{figure*}

\subsection{\textcolor{black}{Ablation Analysis}}
In this part, we will explore multiple factors of our TAD metric, and show their influences on measuring adaptation difficulty for novel tasks.

\textbf{The number of sampled training tasks.}
In previous experiments, we randomly select $n$ different training tasks from training categories.
Subsequently, we calculate the average distance between these selected tasks and a novel task with TAD metric to measure the adaptation difficulty.
As a beginning, we first investigate the impact of the number of sampled training tasks, denoted as $n$, on the measuring of task adaptation difficulty.
We explore different values for $n$ within the set $\{100, 1000, 10000, 100000\}$, and conduct experiments using APNet model on the CUB dataset.
Fig. \ref{fig:sampled tasks} shows the distance-accuracy curves corresponding to different values of $n$.
We can find that, as the task distance increases, the accuracy of APNet generally tends to decrease across various values of $n$.
However, the decreasing tendency is not monotonic with small value of $n$ ($n=100$ and $n=1000$).
When $n=100$ or $1000$, the accuracy of APNet tends to decrease at first and then increase with distance ranging from 0.17 to 0.22.
As the value of $n$ increases (from the left sub-figure to the right one), the decreasing tendency of performance gradually stabilizes, presenting a linear relationship between the few-shot performance and calculated distance.
We argue this is because when a small number of training tasks is sampled (for example, when $n=100$ or $1000$), the sampled tasks can not cover the characteristics of the entire training data. 
Taking CUB dataset as an illustration, CUB dataset consits of 100 training categories.
In the 5-way 1-shot setting, each training task is composed of 5 categories sampled from a pool of 100 training categories.
During the training process, the number of possible sampled training tasks is a combinatorial number, specifically $C_{100}^5=75,287,520$, which is much larger than 100 and 1000.
Our experimental results also indicate that, despite the vast number of possible training tasks, stable decreasing tendency of few-shot performance can be achieved by sampling only $10,000$ or $100,000$ training tasks.

\begin{figure*}
    \centering
    \includegraphics[width=1.0\textwidth]{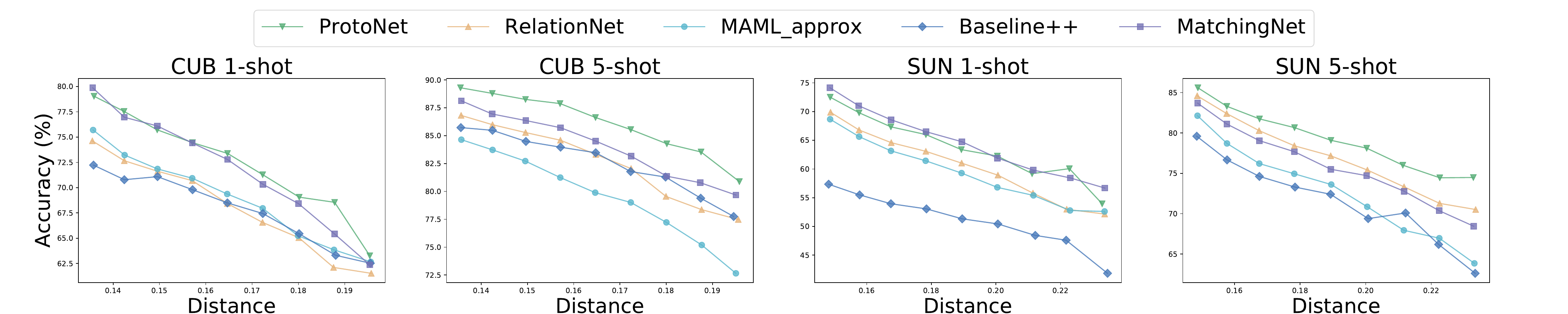}
    \caption{Accuracy of different methods in terms of the average task distance. From left to right, 5-way 1-shot and 5-shot on CUB/SUN. ResNet18 is used as the backbone model. For MAML, we use a first-order approximation in the gradient for memory efficiency (denoted as MAML\_approx). The approximation has been shown to have nearly identical performance as the full version \cite{finn2017maml, chen2019baseline++}.}
    \label{fig:deeper backbone}
\end{figure*}

\textbf{Deeper Backbone.}
As mentioned previous sections, we have proven that TAD can serve as a metric to measure the adaptation difficulty on novel tasks for different FSL methods.
Here we consider how a deeper backbone affects this conclusion.
Following \cite{chen2019baseline++}, we use ResNet18 as backbones and train the five FSL models on CUB and SUN datasets.
Fig. \ref{fig:deeper backbone} shows the task distance and the corresponding accuracy of those models on 2,400 novel tasks.
As shown in Fig. \ref{fig:deeper backbone}, we observe similar phenomenon that with the increase of task distance, the accuracy of these models tends to decrease.
This indicates that the proposed TAD metric still works for different FSL methods with a deeper backbone model.

\begin{figure}
    \centering
    \includegraphics[width=0.5\textwidth]{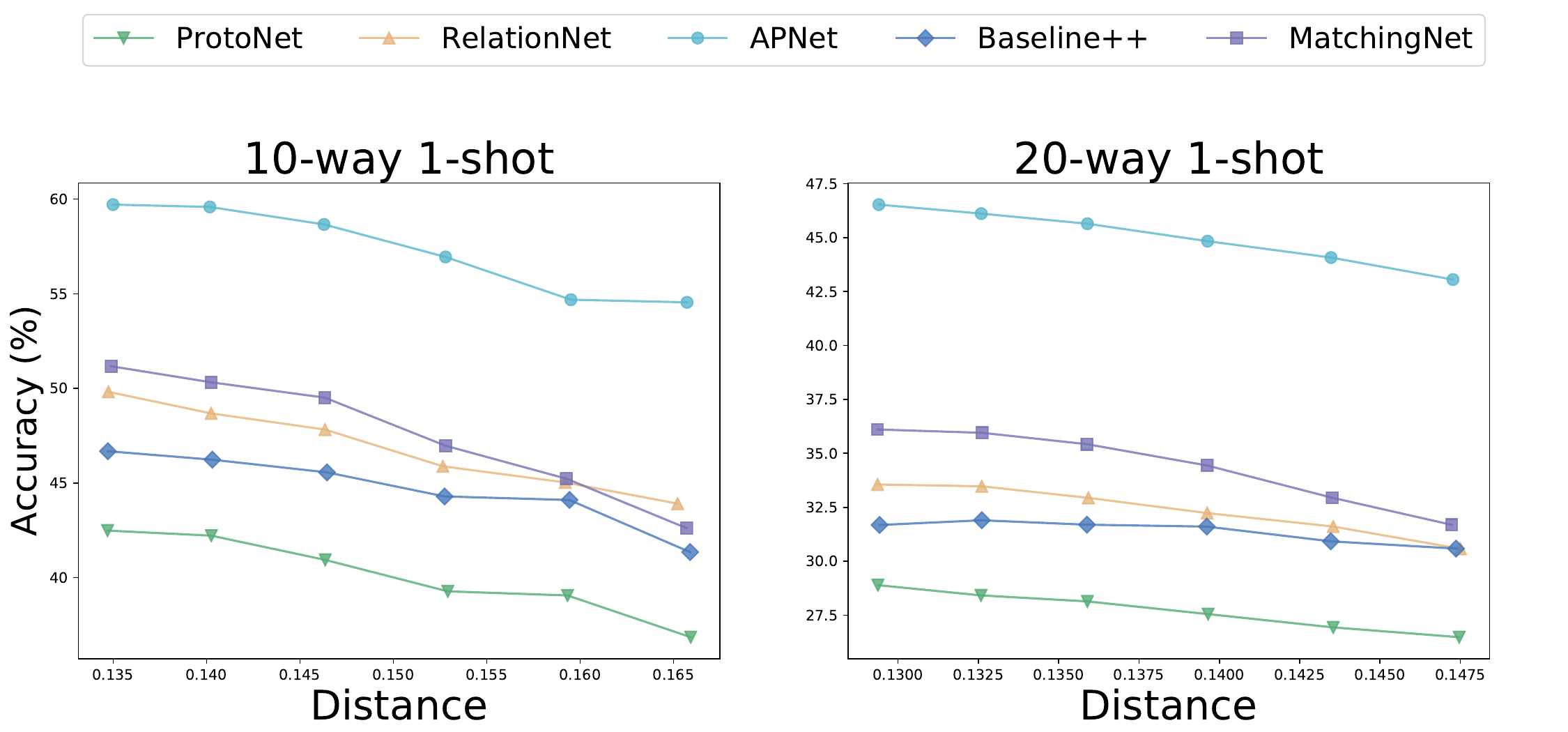}
    \caption{Accuracy of different methods in terms of the average task distance. These models are trained with 5-way 1-shot setting while test with 10-way/20-way 1-shot setting. Note that we have omitted MAML, as it lacks the capacity to deal with varying numbers of ways.}
    \label{fig:varying ways}
\end{figure}

\textbf{Varying number of categories.}
We next investigate the influence of number of categories in novel tasks.
In this exploration, we maintain a constant number of categories in training tasks while varying the number of categories in novel tasks, which involves a discrepancy in the number of categories between the training and novel tasks.
More specifically, we train various FSL models with 5-way 1-shot setting while evaluate them with 10-way/20-way 1-shot setting.
Fig. \ref{fig:varying ways} shows the distance and accuracy curves on CUB and SUN datasets with varying number of categories.
From Fig. \ref{fig:varying ways}, we have two observations: 
(1) TAD metric can still quantify the task relatedness and reflect adaptation difficulty on novel tasks with the discrepancy between training and novel tasks.
Fig. \ref{fig:varying ways} shows a similar phenomenon that with the increase of task distance, the accuracy of different FSL models tends to decrease.
(2) However, TAD metric becomes less applicable to some FSL models as the degree of discrepancy deepens.
For example, when novel tasks contain 20 categories, we observe that the accuracy of Baseline++ model experiences only marginal decline  (around 1\%) with increasing distance.

\subsection{\textcolor{black}{Comparison with Other Metrics}}
Here we present comparisons between proposed TAD and other metrics to show the effectiveness of it.
For comparing different metrics, we design a task selection experiment. 
More specifically, we select top 5\% novel tasks with the highest distances computed by different metrics, and then evaluate the accuracy of FSL models on these chosen tasks. 
The central hypothesis behind this experiment is that if a distance metric can better reflect task difficulty, then novel tasks with the highest distances should be more challenging.

\textbf{Comparative Methods.}
We compare our TAD metric with two distribution divergence metrics, which have been proposed in the few-shot learning or related area: 
(1) Frechet Inception Distance (\textbf{FID}) \cite{milbich2021ooDML}, FID is a metric to measure the distance between two image distributions by comparing their mean and covariance of representations. 
(2) Earth Mover’s Distance (\textbf{EMD}) \cite{oh2022domain-similarity}, EMD is a measure of dissimilarity between two distributions by considering the distance as the cost of moving image representations from one distribution to the other. 
The two above methods assume that the representation function learned from training categories can be directly applied to novel categories.

\textbf{Results.}
We compare these aforementioned methods with both original TAD (defined in Eq. (\ref{eq:origin})) and approximate TAD (defined in Eq. (\ref{eq:approximation})) on CUB and the cross-dataset scenario.
Tab. \ref{tab:comparison metrics cub} and \ref{tab:comparison metrics cross} illustrates the results of different methods. 
First of all, We can find that, with both human-annotated and auto-annotated attributes, original TAD (TAD-Orig) and approximate TAD (TAD-Approx) metrics significantly outperform other three methods in identifying more challenging novel tasks across all FSL models, demonstrating the effectiveness of TAD metrics. 
Secondly, TAD metrics can be applied to all FSL methods, while the FID and EMD metrics does not support MAML.
This is because MAML updates the feature extractor during adaptation, thereby violating the assumption of a common representation function between training and novel categories.
Finally, Tab. \ref{tab:comparison metrics cub} and \ref{tab:comparison metrics cross} also show that approximate TAD metric achieves similar or even better performance than the original one, which showcase that approximating Eq. (\ref{eq:origin}) with Eq. (\ref{eq:approximation}) is still effective and useful in measuring the adaptation difficulty on novel tasks for various FSL models.

\begin{table}[]
\centering
\begin{tabular}{lllll}
\toprule
            & \textbf{FID}        & \textbf{EMD}        & \textbf{TAD-Orig}   & \textbf{TAD-Approx} \\ \midrule
MatchingNet & 2.4 (0.9)    & 2.3 (0.8)    & -6.1 (1.2)        & \textbf{-8.0 (1.0)}           \\
ProtoNet    & 2.6 (0.9)    & -4.4 (0.8)   & -6.7 (0.8)        & \textbf{-6.9 (0.8)}           \\
RelationNet & 2.4 (0.8)    & 1.4 (1.3)    & -7.2 (0.8)        & \textbf{-8.2 (1.0)}           \\
Baseline++  & 4.5 (0.8)    & 4.9 (1.0)    & -4.5 (0.5)        & \textbf{-5.2 (1.0)}           \\
APNet       & 2.2 (0.7)    & 1.7 (0.8)    & -4.6 (1.1)        & \textbf{-7.0 (0.6)}           \\
MAML        & -            & -            & -6.2 (1.1)        & \textbf{-7.3 (1.0)}           \\ \midrule
Mean        & 2.8          & 1.2          & -5.9              & \textbf{-7.0}                 \\ \bottomrule
\end{tabular}
\caption{Comparison with different metrics on CUB dataset.
In the experiment, we run 10 times with different random seed then report the average results with standard deviation. The best results are in bold. The dash indicates that reported results are unavailable.}
\label{tab:comparison metrics cub}
\end{table}

\begin{table}[]
\centering
\begin{tabular}{lllll}
\toprule
            & \textbf{FID}        & \textbf{EMD}        & \textbf{TAD-Orig}   & \textbf{TAD-Approx} \\ \midrule
MatchingNet & 1.4 (0.7)  & 0.8 (1.0)  & \textbf{-3.0 (1.0)} & -2.9 (0.8)  \\
ProtoNet    & 0.7 (0.6)  & -1.1 (0.5) & -1.7 (0.6) & \textbf{-2.2 (0.6)}  \\
RelationNet & 0.7 (0.9)  & 1.0 (0.9)  & \textbf{-4.3 (1.1)} & -4.0 (0.9)  \\
Baseline++  & -0.1 (0.8) & -0.4 (0.7) & \textbf{-3.4 (0.5)} & -2.9 (1.0)  \\ \midrule
Mean        & 0.7        & 0.1        & \textbf{-3.1}       & -3.0        \\ \bottomrule
\end{tabular}
\caption{Comparison with different metrics from miniImageNet to CUB. The best results are in bold.}
\label{tab:comparison metrics cross}
\end{table}

\subsection{\textcolor{black}{Time Complexity}}
We next analyze the time complexity of original and approximate TAD metrics.
Computing the original TAD distance in Eq. (\ref{eq:origin}) requires finding the minimum weight perfect matching $M$ in a bipartite graph $G$, which is a combinatorial optimization problem.
The Hungarian algorithm \cite{kuhn1955hungarian} is commonly used to solve the matching problem, which employs a modified shortest path search in the augmenting path algorithm.
The running time of the Hungarian algorithm is $O(V^2\log V+VE)$, where $V$ is the number of vertices and $E$ is the number of edges in the bipartite graph $G$.
Assume that each training and novel task contain $C$ categories, thus the time complexity of original TAD is $O(C^3)$.
By simplifying the comparison of attribute conditional distribution between two tasks, the approximate TAD can avoid the need to seek the minimum weight perfect matching $M$ and achieve a constant time complexity of $O(1)$.
In Tab. \ref{tab:time complexity}, we compare the computation time of original and approximate TAD metrics on 2,400 novel tasks.
As we can see, the computational efficiency of approximate TAD greatly surpasses the original one, requiring only 0.7 seconds to compute across 2400 novel tasks, underscoring its advantage of ease of computation.
Furthermore, as shown in Tab. \ref{tab:time complexity}, the increase of number of ways (number of categories) in each novel task has significant influence on the computation time of the original TAD, while has negligible influence on the approximate TAD.
This observation validates that the time complexity of approximate TAD is independent of the number of categories in novel tasks.
Finally, the number of shots (number of labeled samples) has negligible influence on the computation time of the two metrics.
This is because both original and approximate TAD metrics rely solely on the attribute conditional distributions in training and novel tasks, making them independent of models and the number of labeled samples in novel tasks.

\begin{table}[!t]
    \centering
    \begin{tabular}{lccl}
    \toprule
    \multicolumn{1}{l}{\textbf{Metric}}   & \multicolumn{1}{l}{\textbf{Ways}} & \multicolumn{1}{l}{\textbf{Shots}} & \textbf{Time}     \\ \midrule
    \multirow{4}{*}{\textbf{Original TAD}}    & 5                        & 1                          & 1728.9 s \\
                             & 5                        & 5                          & 1713.9 s \\ \cmidrule{2-4}
                             & 10                       & 1                          & 1850.8 s \\
                             & 10                       & 5                          & 1859.8 s \\ \midrule
    \multirow{4}{*}{\textbf{Approximate TAD}} & 5                        & 1                          & 0.72 s   \\
                             & 5                        & 5                          & 0.71 s   \\ \cmidrule{2-4}
                             & 10                       & 1                          & 0.74 s   \\
                             & 10                       & 5                          & 0.74 s   \\ \bottomrule
    \end{tabular}
    \caption{Computation time of original and approximate TAD with varying ways and shots of novel tasks. We run the experiments on a single GTX 3090 GPU and only report the computational time of task distance.}
    \label{tab:time complexity}
\end{table}

\section{Applications}
In Sec. \ref{sec:experiments}, we have shown that TAD metric can quantify the task relatedness and measure the adaptation difficulty on novel tasks for various FSL models.
In this section, we shift our focus to the application of TAD metric.
We present two potential applications: data augmentation and test-time intervention based on the measure of task relatedness and task adaptation difficulty, respectively.

\subsection{Data Augmentation}
With the proposed TAD metric, we can effectively quantify the relatedness between training and novel test tasks.
As the training tasks have a sufficient amount of data while the novel test tasks only have a limited number of labeled sample. 
Leveraging the established task relatedness, we can augment the data of novel test tasks by incorporating the information from closely related training tasks.
In this part, we propose a simple prototype calibration method for data augmentation.
We assume that the feature distribution of each category follows a Gaussian distribution, with the mean correlated to the semantic prototype of each category.
With this in mind, the statistics can be transferred from the training tasks to the novel test tasks if we acquire how related the two tasks are.
To acquire such statistics, we compute the mean of feature vectors for all available labeled samples belonging to the same category.
Subsequently, we calculate the average distance between each novel task and 100,000 training tasks, as done in previous experiments.
Based on these established distances, we remove most unrelated training tasks, and only keep the $K$ most related tasks.
To obtain the transferable statistics from $K$ most related training tasks, we utilize the bipartite graph matching to match each category in a training and novel task.
To this end, we represent the set of categories in a training task and the set of categories in novel task as two disjoint vertex sets respectively, and construct a weighted bipartite graph $G$ between the two vertex sets, where each node denotes a
category and each edge weight represents the distance between
two categories in training and novel task, respectively.
After that, we average the prototypes of matched training categories and integrate them into each support sample feature of the corresponding novel category.

\textbf{Details. } 
We conduct experiments on the CUB and SUN datasets with 5-way 1-shot setting.
In the prototype calibration process, we select $K = 200$ related training tasks based on calculated distances. 
Moreover, to reduce the impact of noise, we implement a filtering mechanism for each novel category.
Specifically, we retain 5 training categories that exhibited the highest frequency during the matching process, thereby refining the prototype calibration and reducing the influence of irrelevant information.

\begin{table}[!t]
\centering
\begin{tabular}{llll}
\toprule
\textbf{Method}                       & \textbf{Method}       & \textbf{CUB}        & \textbf{SUN}        \\ \midrule
\multirow{4}{*}{ProtoNet \cite{snell2017prototypical}}    & Raw          & 57.0       & 60.2       \\ \cmidrule{2-4} 
                             & +TAD ($K=1$)   & 60.3 $\uparrow$ 3.3 & 61.7  $\uparrow$ 1.5 \\
                             & +TAD ($K=10$)  & 63.6 $\uparrow$  6.6 & 65.2 $\uparrow$  5.0 \\
                             & +TAD ($K=200$) & 64.8 $\uparrow$  7.8 & 67.1  $\uparrow$ 6.9 \\ \midrule
\multirow{4}{*}{RelationNet \cite{vinyals2016matchnet}} & Raw          & 61.9       & 60.5       \\ \cmidrule{2-4} 
                             & +TAD ($K=1$)   & 63.7 $\uparrow$  1.6 & 63.5 $\uparrow$  3.0 \\
                             & +TAD ($K=10$)  & 67.7 $\uparrow$ 5.8 & 66.9 $\uparrow$  6.4 \\
                             & +TAD ($K=200$) & 68.7 $\uparrow$  6.8 & 68.2 $\uparrow$  7.7 \\ \bottomrule
\end{tabular}
\caption{Data augmentation experiments on the CUB and SUN datasets. Our proposed TAD metric can effectively improve the performance of metric-based FSL models. The relative  improvements over the baselines are indicated ($\uparrow$).}
\label{tab:application data augmentation}
\end{table}

\textbf{Results. }
The comparison between baselines and our methods are shown in Tab. \ref{tab:application data augmentation}.
As we can see, our TAD metric can effectively improve the performance of metric-based FSL models.
Moreover, with the increase of $K$, the performance improvement of data augmentation becomes significant.
This results indicate that the inaccurate feature distributions of novel tasks can be gradually calibrated by more information from closely related training tasks.

\begin{table}[!t]
\centering
\begin{tabular}{llll}
\toprule
\textbf{Method  }                     & \textbf{Method}             & \textbf{$Acc_{5}$}     & \textbf{$Acc_{10}$}    \\ \midrule
\multirow{4}{*}{ProtoNet}    & Raw                & 47.1     & 48.7     \\ \cmidrule{2-4} 
                             & + TAD (Imbalanced) & 47.6 $\uparrow$ 0.5 & 49.8 $\uparrow$ 1.1 \\
                             & + TAD (Balanced)   & 48.7 $\uparrow$ 1.6 & 50.7 $\uparrow$ 2.0 \\
                             & + GT               & 50.9 $\uparrow$ 3.8 & 52.8 $\uparrow$ 4.1 \\ \midrule
\multirow{4}{*}{RelationNet} & Raw                & 47.5     & 50.0     \\ \cmidrule{2-4} 
                             & + TAD (Imbalanced) & 49.7 $\uparrow$ 2.2 & 51.4 $\uparrow$ 1.4 \\
                             & + TAD (Balanced)   & 50.0 $\uparrow$ 2.5 & 51.7 $\uparrow$ 1.7 \\
                             & + GT               & 53.1 $\uparrow$ 5.6 & 55.2 $\uparrow$ 5.2 \\ \midrule
\multirow{4}{*}{MAML}        & Raw                & 46.7     & 48.6     \\ \cmidrule{2-4} 
                             & + TAD (Imbalanced) &  41.2 $\downarrow$ 5.5        & 43.6 $\downarrow$ 5.0         \\
                             & + TAD (Balanced)   & 49.1 $\uparrow$ 2.4 & 50.7 $\uparrow$ 2.1 \\
                             & + GT               & 51.4 $\uparrow$ 4.7 & 53.8 $\uparrow$ 5.2 \\ \midrule
\multirow{4}{*}{Baseline++}  & Raw                & 49.5     & 52.2     \\\cmidrule{2-4}
                             & + TAD (Imbalanced) & 51.3 $\uparrow$ 1.8       &  53.3 $\uparrow$ 1.1        \\
                             & + TAD (Balanced)   & 51.4 $\uparrow$ 1.9 & 53.6 $\uparrow$ 1.4         \\
                             & + GT               & 54.0 $\uparrow$ 4.5 & 57.1 $\uparrow$ 4.9 \\ \bottomrule
\end{tabular}
\caption{Test-time intervention experiments on CUB dataset. The GT denotes the Ground-Truth method that we directly remove these worst tasks according to its accuracy, which can be seen as an upper bound method. The relative improvements over the baselines are indicated ($\uparrow$).}
\label{tab:test-time intervention}
\end{table}

\subsection{Test-time Intervention}
With the proposed TAD metric, we can measure the adaptation difficulty of novel test tasks without training and adapting a model. This makes sense in real-world scenarios, as we can identify more challenging tasks before training and take interventions to improve a model’s performance on those tasks. In this part, we explore a simple test-time intervention operation that supplies more labeled samples for harder novel tasks. To construct the intervention experiment, we first calculate the distances between each novel task and 100,000 training tasks, as done in previous experiments. We manually set a constant threshold value, denoted as $r$, to identify the harder tasks that exhibit large distances and generally yield low accuracy. Once we identify these tasks, we intervene by providing additional labeled samples, which contain more task-specific information. 

\textbf{Details. }
We run experiments on CUB with 5-way 5-shot setting.
We set the distance threshold $r$ = 0.18 for the CUB dataset. With the above threshold value, we only need to intervene with a small subset of novel tasks (approximately 5\%-7\%), which greatly reduces the cost of sample annotation.
For test-time intervention, we consider two strategies: (1) Balanced Intervention (\emph{Balanced}): We offer 25 extra labeled samples for each intervened task (5 labeled samples for each of the 5 categories), which can be seen as the additional annotated samples. (2) Imbalanced Intervention (\emph{Imbalanced}): In this strategy, we randomly select 25 unlabeled samples from an unlabeled sample pool for each intervened task.
The ground-truth labels for these selected samples are obtained using an oracle. 
Note that, unlike the Balanced Intervention, the samples chosen in the Imbalanced Intervention may exhibit class imbalance since they are randomly selected from a pool.
Following previous work \cite{fu2022worst}, we report the average accuracy of Top-$K$ worst novel tasks, namely $Acc_{K}$, to evaluate the effectiveness of test-time intervention methods.
Specifically, we report the average accuracy of the worst 5 (\textbf{$Acc_{5}$}) and worst 10 (\textbf{$Acc_{10}$}) tasks among the whole 2,400 novel tasks. 

\textbf{Results. }
The quantitative results of the test-time intervention are presented in Tab. \ref{tab:test-time intervention}. As illustrated in Tab. \ref{tab:test-time intervention}, our proposed TAD metric can significantly improve the worst-task performance of different FSL models. 
Furthermore, Tab. \ref{tab:test-time intervention} reveals that imbalanced intervention method is not always effective across various FSL models.
For example, for MAML, the imbalanced intervention method performs even worse than the baseline in term of $Acc_{5}$ and $Acc_{10}$, with a deviation of around 5\%, significantly underperforming balanced intervention and Ground Truth (GT) methods.
This disparity is attributed to MAML's optimization strategy, which requires optimizing all parameters and makes it more susceptible to imbalances in the training data.

\begin{figure}
    \centering
    \includegraphics[width=0.4\textwidth]{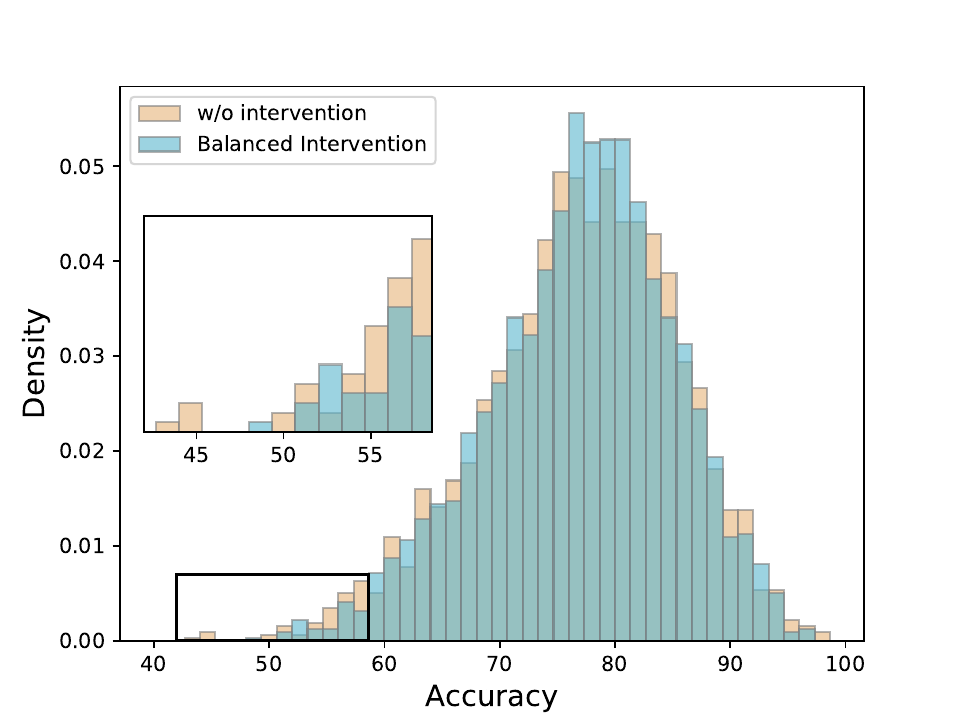}
    \caption{Comparison of task accuracy distributions before and after intervention. The low task accuracy instances are enclosed with a rectangle and enlarged in the middle of figure.}
    \label{fig:test-time distribution}
\end{figure}

To further understand how the test-time intervention method improves the worst-task performance of FSL models, 
we compare the distributions of task accuracy before and after test-time intervention.
In Fig. \ref{fig:test-time distribution}, we illustrate this comparison using the ProtoNet model and the CUB dataset. 
Notably, our analysis reveals that, with a balanced intervention approach, there is a notable decrease in the density of low task accuracy instances (enclosed within a rectangle). 
For example, before intervention, certain challenging novel tasks exist where the ProtoNet model achieves only 45\% accuracy. 
Yet, following the application of a balanced intervention, these challenging tasks are successfully addressed.
Simultaneously, there is a discernible increase in the density of high task accuracy instances (the intervals of accuracy around 80\%) due to the intervention.
These results underscore the efficacy of the TAD metric in identifying more challenging tasks.
Furthermore, the implementation of test-time intervention proves to be pivotal in enhancing the worst-task performance, and all of this is achieved with minimal additional annotation costs, requiring intervention in only 5\%-7\% of novel tasks.

\section{Conclusion}
We propose a novel distance metric, called Task Attribute Distance (TAD), to quantify the relationship between training and novel tasks in FSL, which is build on the category-level attribute annotations.
We present theoretical analysis of the generalization error bound on a novel task with TAD, which connects task relatedness and adaptation difficulty theoretically.
Our experiments demonstrate TAD can effectively reflect the task adaptation difficulty for various FSL methods, even if some of them do not learn attributes explicitly or human-annotated attributes are not available.
We further present two potential applications of  TAD metric: selecting the most related training tasks for data augmentation and intervening the challenging novel test tasks to improve the worst-task performance of FSL models.
We believe our theoretical and empirical analysis can provide more insight into few-shot learning and related areas.

\textbf{Limitations and broader impact.}
Our theoretical and empirical analysis on the  proposed TAD metric assume that attributes are conditionally independent without considering the correlations that may exist between them. 
Furthermore, identifying which attributes are critical in the distance metric is still an open question. 
However, we believe that our work lays a solid foundation for measuring task relatedness and adaptation difficulty of novel tasks, which offers a starting point for further research in FSL and related areas.

\section*{Acknowledgments}
This work is partially supported by National Key R$\&$D Program of China no. 2021ZD0111901, and
National Natural Science Foundation of China (NSFC): 62376259 and 62276246.



{{
      \bibliographystyle{IEEEtran}
      \bibliography{overleaf/bibtex_v2310/bilevel_optimization, overleaf/bibtex_v2310/domain_adaptation, overleaf/bibtex_v2310/few-shot_learning, overleaf/bibtex_v2310/graph_neural_networks, overleaf/bibtex_v2310/others, overleaf/bibtex_v2310/transfer_learning, overleaf/bibtex_v2310/scene_graph, overleaf/bibtex_v2310/theory}

}}

\end{document}


\title{\emph{Appendix of} \\
Task Attribute Distance for Few-Shot Learning: Theoretical Analysis and Applications}

\maketitle

\appendices

\section{Theoretical Results and Proofs}
Recall that we have defined the distance between two categories $y_k, y_t$ in Section III.B of the main paper,
expressed as $d_{\mathcal{A}}(y_k, y_t) = \sum_{a_i\in\mathcal{A}}\frac{1}{2L}\sum_{l=1}^L\left|p(a_i^l|y_k)-p(a_i^l|y_t)\right|$ if attribute space $\mathcal{A}$ is countable.
Based on the distance definition, we introduce the following theoretical results and proofs.

\begin{lemma}
    Let $\mathcal{A}$ be the attribute space, $L$ be the number of attributes.
    Assume all attributes are independent of each other given the class label, i.e. $p(a|y) = \prod_{l=1}^Lp(a^l|y)$.
    For all $a_i \in \mathcal{A}$ and any two categories $y_k, y_t$, the following inequality holds:
    \begin{align}
        \sum_{a_i\in\mathcal{A}}\left|p(a_i|y_k)-p(a_i|y_t)\right| \leq d_{\mathcal{A}}(y_k, y_t) +  \Delta,
        \label{eq:distance ineq}
    \end{align}
    where $\Delta=\sum_{a_i\in\mathcal{A}}\frac{1}{2L}\sum_{l=1}^L(p(a^l_i|y_k)+p(a^l_i|y_t))$.
    \label{lemma 1}
\end{lemma}
\begin{proof}
    Firstly, the following inequality holds,
    \begin{equation}
        \frac{1}{L}\sum_{i=1}^L x_i \geq (\prod_{i=1}^L x_i)^{\frac{1}{L}} \geq \prod_{i=1}^L x_i,
        \label{eq:AM-GM}
    \end{equation}
    where $x_i$ is a non-negative real number and ranges from 0 to 1.
    The proof of Eq. (\ref{eq:AM-GM}) is straightforward: the first inequality is an application of AM-GM inequality (or the inequality of arithmetic and geometric means) on a list of $L$ non-negative real numbers $\{x_1,...,x_L\}$, and the second inequality holds because $\prod_{i=1}^L x_i$ ranges from $0$ to $1$.
    
    Next, we try to prove Eq. (\ref{eq:distance ineq}) based on the above inequality.
    If all attributes are independent of each other given the class label, for any conditional probabilities $p(a_i|y_k)$ and $p(a_i|y_t)$, we have  
    \begin{align}
        \left|p(a_i|y_k)-p(a_i|y_t)\right| &\leq \max(p(a_i|y_k), p(a_i|y_t)) \notag \\
        &= \max(\prod_{l=1}^L p(a_i^l|y_k), \prod_{l=1}^L p(a_i^l|y_t)).
    \end{align}
    Note that $p(a_i^l|y_k)$ and $p(a_i^l|y_t)$ are both  real numbers between 0 and 1.
    Thus, combining with Eq. (\ref{eq:AM-GM}), we have
    \begin{align}
        \left|p(a_i|y_k)-p(a_i|y_t)\right| &\leq \max (\frac{1}{L}\sum_{l=1}^Lp(a_i^l|y_k), \frac{1}{L}\sum_{l=1}^Lp(a_i^l|y_t)) \notag \\
        &\leq \frac{1}{L} \sum_{l=1}^L\max (p(a_i^l|y_k), p(a_i^l|y_t)) \notag \\
        &\leq \frac{1}{2L} \sum_{l=1}^L (\left|p(a_i^l|y_k)-p(a_i^l|y_t)\right|) + \frac{1}{2L} \sum_{l=1}^L(p(a_i^l|y_k)+p(a_i^l|y_t)).
    \end{align}
    For all $a_i\in\mathcal{A}$, denote $\Delta=\sum_{a_i\in\mathcal{A}}\frac{1}{2L}\sum_{l=1}^L(p(a^l_i|y_k)+p(a^l_i|y_t))$, we have
    \begin{align}
        \sum_{a_i\in\mathcal{A}}\left|p(a_i|y_k)-p(a_i|y_t)\right| \leq d_{\mathcal{A}}(y_k, y_t) +  \Delta.
    \end{align}
\end{proof}

\begin{theorem}
    With the same notation and assumptions as Lemma 1, 
    let $\mathcal{H}$ be the hypothesis space with VC-dimension $d$, 
    $f_\theta$ and $g_\phi$ be the embedding function and prediction function as introduced in Section IV.A respectively. 
    Denote $g_{\phi^*}$ as the best prediction function on some specific tasks given a learned embedding function.
    For any training task $\tau_i=(\mathcal{D}_i, S_i)$ and novel task $\tau_{j}'=(\mathcal{D}_{j}', S_{j}')$, suppose the number of categories in the two tasks is the same, then with probability at least $1-\delta$, $\forall g_\phi\circ f_\theta\in\mathcal{H}$, we have 
     \begin{align}
            \epsilon(\theta, \phi; \tau_{j}')&\leq 
            \hat{\epsilon}(\theta, \phi; \tau_i) + \sqrt{\frac{4}{m_i}(d\log\frac{2em_i}{d}+\log\frac{4}{\delta})} + d_\theta(\tau_i, \tau_{j}') + \Delta' + \lambda,     
    \end{align}
    where $\lambda=\lambda_i+\lambda_j'$ is the generalization error of $g_{\phi^*}$ on the two tasks, i.e., 
    $\lambda_i=\mathbb{E}_{(x,y)\sim \mathcal{D}_i}[\mathbb{I}(g_{\phi_{i}^*}(f_{\theta_i}(x))\neq y)]$, $\lambda_j'=\mathbb{E}_{(x,y)\sim \mathcal{D}_{j}'}[\mathbb{I}(g_{\phi_{j}'^*}(f_{\theta_{j}'}(x))\neq y)]$. $\Delta'$ is a term depending on learned prediction functions  $g_{\phi_i}, g_{\phi_j'}$ and the best prediction functions $g_{\phi_i^*}, g_{\phi_j'^*}$.
    \label{theorem 1}
\end{theorem}
\begin{proof}
    Note that the error $\epsilon(\theta, \phi; \tau_{j}')=\mathbb{E}_{(x,y)\sim \mathcal{D}_{j}'}[\mathbb{I}(g_{\phi_{j}'}(f_{\theta_{j}'}(x))\neq y)]$ can be decomposed into two parts: with the same $f_{\theta_{j}'}$, (1) the probability that the learned prediction function $g_{\phi_j'}$ agrees with the best prediction function $g_{\phi_{j}'^*}$, but they both output the wrong prediction; and (2) the probability that the learned prediction function $g_{\phi_j'}$ disagrees with the best prediction function $g_{\phi_{j}'^*}$, while $g_{\phi_j'}$ outputs the wrong prediction.
    The first part can be bounded by the error of $g_{\phi_{j}'^*}$, and the second part can be bounded by the probability that $g_{\phi_j'}$ disagrees with $g_{\phi_{j}'^*}$.
    Denote $Z^{j'}=\{(x,y)|g_{\phi_j'}(f_{\theta_{j}'}(x))\neq g_{\phi_{j}'^*}(f_{\theta_{j}'}(x)), (x,y)\sim \mathcal{D}_j'\}$, then $P_{\mathcal{D}_j'}[Z^{j'}]$ represents the probability that $g_{\phi_j'}$ disagrees with $g_{\phi_j'^*}$ based on the same embedding function $f_{\theta_{j}'}$ on distribution $\mathcal{D}_j'$.
    We have
    \begin{align}
        \epsilon(\theta, \phi; \tau_{j}') &\leq \lambda_j' + P_{\mathcal{D}_j'}[Z^{j'}] \notag \\
        &= \lambda_j' +  P_{\mathcal{D}_i}[Z^i] + P_{D_j'}[Z^{j'}] - P_{\mathcal{D}_i}[Z^i] \notag \\
        &\leq  \lambda_j' + \lambda_i + \epsilon(\theta, \phi; \tau_i) + P_{\mathcal{D}_j'}[Z^{j'}] - P_{D_i}[Z^i] \notag \\
        &= \lambda + \epsilon(\theta, \phi; \tau_i) + P_{\mathcal{D}_j'}[Z^{j'}] - P_{\mathcal{D}_i}[Z^i].
    \end{align}
    Assume that the two tasks both have $C$ categories and $p(y)$ is uniform, we can decompose the probability $P_{\mathcal{D}_j'}[Z^{j'}]$ as $P_{\mathcal{D}_j'}[Z^{j'}] = \frac{1}{C}\sum_{t=1}^C P_{x|y_t}[Z^{j'}_t]$, where $Z^{j'}_t = \{x|g_{\phi_j'}(f_{\theta_{j}'}(x))\neq g_{\phi_j'^*}(f_{\theta_{j}'}(x)), x\sim p(x|y_t)\}$.
    Thus, we have
    \begin{align}
        \epsilon(\theta, \phi; \tau_{j}') &\leq \lambda + \epsilon(\theta, \phi; \tau_i) + \frac{1}{C}(\sum_{t=1}^C P_{x|y_t}[Z^{j'}_t] - \sum_{k=1}^C P_{x|y_k}[Z^i_k]) \notag \\
        &= \lambda + \epsilon(\theta, \phi; \tau_i) + \frac{1}{C}\sum_{e_{tk}\in M}(P_{x|y_t}[Z^{j'}_t] - P_{x|y_k}[Z^i_k]),
        \label{eq:uninduced}
    \end{align}
    where $M$ is a maximum matching, which contains $C$ edges and each edge 
    $e_{tk}\in M$ links two categories $y_t, y_k$ in task $\tau_j'$ and $\tau_i$ respectively.
    
    Next, we consider to replace the conditional distribution $p(x|y)$ with the attribute conditional distribution $p(a|y)$, because the former is usually unknown and difficult to estimate.
    For a conditional distribution $p(x|y)$ and a mapping $f_{\theta_{j}'}:\mathcal{X}\rightarrow \mathcal{A}$, a new distribution can be induced over the space $\mathcal{A}$ as $p_{\theta_{j}'}(a|y)\triangleq p(f_{\theta_{j}'}(x)|y)$.
    Based on the induced distribution $p_{\theta_{j}'}(a|y)$, we have $P_{x|y_t}[Z^{j'}_t]=P_{a|y_t}[\{a|g_{\phi_j'}(a)\neq g_{\phi_j'^*}(a), a\sim p_{\theta_{j}'}(a|y_t)\}]$.
    For clarity, we define $A_t=\{a|g_{\phi_j'}(a)\neq g_{\phi_j'^*}(a), a\sim p_{\theta_{j}'}(a|y_t) \}$ and $A_k=\{a|g_{\phi_i}(a)\neq g_{\phi_i^*}(a), a\sim p_{\theta_i}(a|y_k) \}$.
    Thus, Eq. (\ref{eq:uninduced}) can be rewritten as
    \begin{align}
        \epsilon(\theta, \phi; \tau_j') &\leq \lambda + \epsilon(\theta, \phi; \tau_i) + \frac{1}{C}\sum_{e_{tk}\in M}(P_{a|y_t}[A_t] - P_{a|y_k}[A_k]).
        \label{eq: replace distribution}
    \end{align}
    
    Let $A_{t\cup k}=A_t\cup A_k$ be the union set of $A_t$ and $A_k$, $A_{t\cap k}=A_t\cap A_k$ be the intersection set of $A_t$ and $A_k$, then we have another inequality as
    \begin{align}
        P_{a|y_t}[A_t] - P_{a|y_k}[A_k]
        &\leq(P_{a|y_t}[A_{t\cup k}] - P_{a|y_k}[A_{t\cup k}]) + \left|P_{a|y_t}[A_{t\cap k}] - P_{a|y_k}[A_{t\cap k}]\right| \notag \\
        &+ \left|P_{a|y_t}[A_k] - P_{a|y_k}[A_t]\right|.
        \label{eq:set decomposition}
    \end{align}
    For clarity, we use two notions $\Delta_1$ and $\Delta_2$ to denote $\frac{1}{C}\sum_{e_{tk}\in M}\left|P_{a|y_t}[A_{t\cap k}] - P_{a|y_k}[A_{t\cap k}]\right|$ and $\frac{1}{C}\sum_{e_{tk}\in M}\left|P_{a|y_t}[A_k] - P_{a|y_k}[A_t]\right|$, respectively.
    Based on Lemma 1 and Eq. (\ref{eq:set decomposition}), we have
    \begin{align}
        \frac{1}{C}\sum_{e_{tk}\in M}(P_{a|y_t}[A_t] - P_{a|y_k}[A_k]) &\leq
        \frac{1}{C}\sum_{e_{tk}\in M}(P_{a|y_t}[A_{t\cup k}] - P_{a|y_k}[A_{t\cup k}]) + \Delta_1 + \Delta_2 \notag \\
        &= \frac{1}{C}\sum_{e_{tk}\in M}\sum_{a_i\in A_{t\cup k}}(p_{\theta_i}(a_i|y_t) - p_{\theta_j'}(a_i|y_k))  + \Delta_1 + \Delta_2 \notag \\
        &\leq \frac{1}{C}\sum_{e_{tk}\in M}d_\theta(y_t, y_k) + \Delta  + \Delta_1 + \Delta_2 \notag \\
        &= d_\theta(\tau_j', \tau_i) + \Delta  + \Delta_1 + \Delta_2,
        \label{eq: apply lemma1}
    \end{align}
    where $\Delta=\frac{1}{C}\sum_{e_{tk}\in M}\sum_{a_i\in A_{t\cup k}}\frac{1}{2L}\sum_{l=1}^L (p_{\theta_i}(a_i^l|y_k)+p_{\theta_j'}(a_i^l|y_t))$.
    Denoting $\Delta' = \Delta + \Delta_1 + \Delta_2$, and combining Eq. (\ref{eq: replace distribution}) and Eq. (\ref{eq: apply lemma1}), we can get
    \begin{align}
        \epsilon(\theta, \phi; \tau_j') 
        &\leq \lambda + \epsilon(\theta, \phi; \tau_i) + d_\theta(\tau_i, \tau_j') + \Delta'.
        \label{eq:wo_vc}
    \end{align}

    Finally, we apply Vanik-Chervonenkis theory \cite{vapnik1999vc-theory} to bound the generalization error $\epsilon(\theta, \phi; \tau_i)$ in Eq. (\ref{eq:wo_vc}) by its empirical estimate $\hat{\epsilon}(\theta, \phi; \tau_i)$.
    Namely, if $S_i$ is a $m_i$-size i.i.d sample set, then with probability at least $1-\delta$,
    \begin{equation}
        \epsilon(\theta, \phi; \tau_i) \leq \hat{\epsilon}(\theta, \phi; \tau_i) + \sqrt{\frac{4}{m_i}(d\log\frac{2em_i}{d}+\log\frac{4}{\delta})}.
        \label{eq:vc}
    \end{equation}
    Combining with Eq. (\ref{eq:wo_vc}),  with probability at least $1-\delta$, we have
    \begin{align}
        \epsilon(\theta, \phi; \tau_{j}')&\leq 
            \hat{\epsilon}(\theta, \phi; \tau_i) + \sqrt{\frac{4}{m_i}(d\log\frac{2em_i}{d}+\log\frac{4}{\delta})} + d_\theta(\tau_i, \tau_{j}') + \Delta' + \lambda.
    \end{align}
\end{proof}

\begin{corollary}

   With the same notation and assumptions as Theorem 1, 
    for $n$ training tasks $\{\tau_i\}_{i=1}^n$ and a novel task $\tau_{j}'$, 
    define $\hat{\epsilon}(\theta, \phi; \tau_{i=1}^n)=\frac{1}{n}\sum_{i=1}^n \hat{\epsilon}(\theta, \phi; \tau_i)$, then with probability at least $1-\delta$, $\forall g_\phi\circ f_\theta\in\mathcal{H}$, we have
    \begin{align}
            \epsilon(\theta, \phi; \tau_{j}')&\leq 
            \hat{\epsilon}(\theta, \phi; \tau_{i=1}^n) + \frac{1}{n}\sum_{i=1}^n\sqrt{\frac{4}{m_i}(d\log\frac{2em_i}{d}+\log\frac{4}{\delta})} + \frac{1}{n}\sum_{i=1}^n d_\theta(\tau_i, \tau_{j}') + \Delta' + \lambda,  
    \end{align}
    where $\lambda=\frac{1}{n}\sum_{i=1}^n \lambda_i + \lambda_j'$, and $\Delta'$ is a term depending on the learned prediction functions $\{g_{\phi_i}\}_{i=1}^n, g_{\phi_j'
    }$ and the best prediction functions $\{g_{\phi_i^*}\}_{i=1}^n, g_{\phi_j'^*}$.
    \label{corollary 1}
\end{corollary}
\begin{proof}
    The proof of \textbf{Corollary 1} is similar to the proof of \textbf{Theorem 1}.
    Denote $\lambda=\frac{1}{n}\sum_{i=1}^n \lambda_i + \lambda_j'$, 
    we have
    \begin{align}
        \epsilon(\theta, \phi; \tau_j') &\leq \lambda_j' + P_{\mathcal{D}_j'}[Z^{j'}] \notag \\
        &= \lambda_j' +  \frac{1}{n}\sum_{i=1}^n P_{\mathcal{D}_i}[Z^i] + P_{D_j'}[Z^{j'}] - \frac{1}{n}\sum_{i=1}^n P_{\mathcal{D}_i}[Z^i] \notag \\
        &\leq  \lambda_j' + \frac{1}{n}\sum_{i=1}^n\lambda_i + \epsilon(\theta, \phi; \tau_{i=1}^n) + P_{\mathcal{D}_j'}[Z^{j'}] - \frac{1}{n}\sum_{i=1}^n P_{D_i}[Z^i] \notag \\
        &= \lambda + \epsilon(\theta, \phi; \tau_{i=1}^n) + \frac{1}{n}\sum_{i=1}^n (P_{\mathcal{D}_j'}[Z^{j'}] - P_{\mathcal{D}_i}[Z^i]).
    \end{align}
    Now, we can follow the same procedure as the proof in \textbf{Theorem 1} and have the following inequlaity
    \begin{align}
        \epsilon(\theta, \phi; \tau_j') &\leq \lambda + \hat{\epsilon}(\theta, \phi; \tau_{i=1}^n) + \frac{1}{n}\sum_{i=1}^n\sqrt{\frac{4}{m_i}(d\log\frac{2em_i}{d}+\log\frac{4}{\delta})} + \frac{1}{n}\sum_{i=1}^n d_\theta(\tau_i, \tau_{j}') + \Delta',
    \end{align}
    where $\Delta' = \frac{1}{n}\sum_{i=1}^n\Delta_i'$, and $\Delta_i'$ corresponds to the additional non-negative term as in Eq. (\ref{eq:wo_vc}), which is derived from $P_{\mathcal{D}_j'}[Z^{j'}] - P_{\mathcal{D}_i}[Z^i]$.
\end{proof}

\begin{theorem}
   \label{theorem 2}
     With the same notation and assumptions as in Corollary \ref{corollary 1}, assume that the conditional distribution $p(x|a^l)$ is task agnostic, 
    \textcolor{black}{and the embedding function $f_{\theta}$ is a $\xi$-approximation meta mapping for $n$ training tasks.
    If $\xi$ tends to zero, the following equality holds:}
    \begin{align}
            \frac{1}{n}\sum_{i=1}^nd_\theta(\tau_i, \tau_{j}')&\leq\frac{1}{n}\sum_{i=1}^nd(\tau_i, \tau_{j}').
    \end{align}
\end{theorem}
\begin{proof}
    Without loss of generality, we first consider a single training task $\tau_i$ and prove $d_\theta(\tau_i, \tau_{j}')\leq d(\tau_i, \tau_{j}')$.
    Assume the embedding function $f_{\theta}$ is a $\xi$-approximation meta mapping for $n$ training tasks and $\xi$ tends to zero.
    Thus, for any training task $\tau_i$, the induced distribution $p_{\theta_i}(a|y_k)$ is equal to the ground-truth distribution $p(a|y_k)$ for task $\tau_i$.
    Thus, we have $d_\theta(\tau_i, \tau_j')=\frac{1}{C}\sum_{e_{kt}\in M} \frac{1}{L}\sum_{l=1}^{L} d_{L_1}(p(a^l|y_k), p_{\theta_{j}'}(a^l|y_t))$, which measures the distance between the ground-truth distribution $p(a^l|y_k)$ for task $\tau_i$ and the induced distribution $p_{\theta_{j}'}(a^l|y_t)$ for task $\tau_j'$.
    Next, for any attribute $l$, we consider three cases:
    (1) the values of attribute $l$ in training and novel tasks are disjoint, which means it is a new attribute or new values of observed attribute for novel task $\tau_j'$.
    In this case, for any two categories $y_k$ and $y_t$, the model-related distance $d_{L_1}(p(a^l|y_k), p_{\theta_j'}(a^l|y_t)) \leq d_{L_1}(p(a^l|y_k), p(a^l|y_t))=1$;
    (2) the values of attribute $l$ in training and novel tasks are completely overlapped.
    As the conditional distribution $p(x|a^l)$ is task-agnostic, the attribute classifier $f_{\theta_j'}$ can also identify attribute $l$ in novel task $\tau_{j}'$.
    In this case, $d_{L_1}(p(a^l|y_k), p_{\theta_j'}(a^l|y_t)) = d_{L_1}(p(a^l|y_k), p(a^l|y_t))$;
    (3) the values of attribute $l$ in training and novel tasks are overlapped but not the same.
    We can divide the values of attribute $l$ into two parts: the completely overlapped values and the disjoint values, then we can follow the same analysis procedures as in the case (1) and case (2).

    In summary, we can arrive that the model-related distance $d_\theta(\tau_i, \tau_j')$ is no more than the model-agnostic distance $d(\tau_i, \tau_j')$, thus we have $\frac{1}{n}\sum_{i=1}^nd_\theta(\tau_i, \tau_{j}')\leq\frac{1}{n}\sum_{i=1}^nd(\tau_i, \tau_{j}')$.
    
\end{proof}

\section{Attribute Prototypical Network}
\begin{figure}[t]
  \centering
  \includegraphics[width=0.7\linewidth]{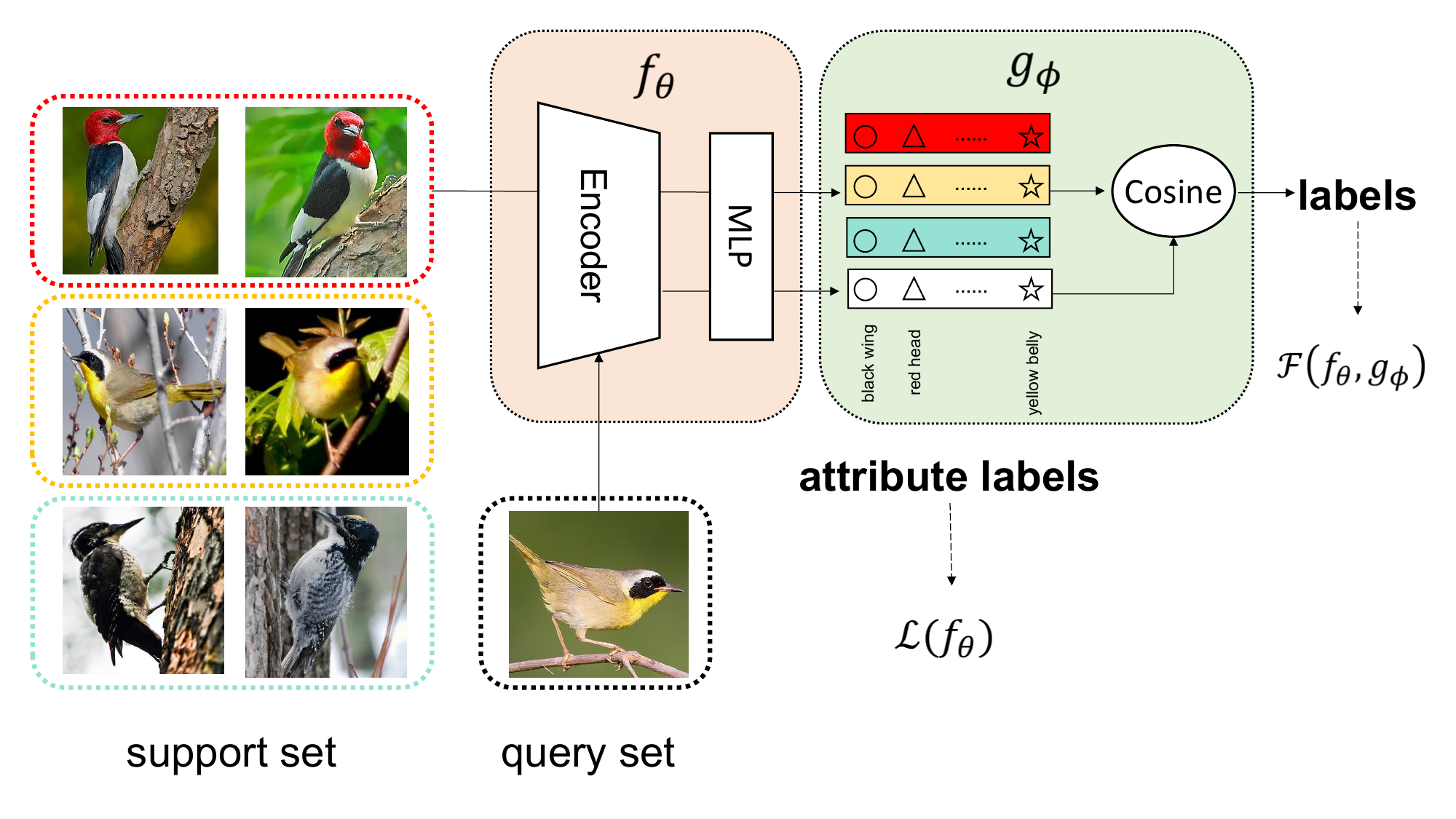}
   \caption{A sketch of APNet.}
   \label{fig:apnet}
\end{figure}

Our theoretical analysis is based on a specific meta-learning framework with attribute learning.
Thus, we instantiate a simple model under that framework as an example, we call this model Attribute Prototypical Network (APNet).
A sketch of APNet is presented in Eq. (\ref{fig:apnet}).

Let $S=\{(x_k,y_k)\}_{k=1}^m$ include all labeled samples in $n$ training tasks. 
For each sample $(x_k, y_k)\in S$, assume we have $L$ binary 
attribute labels $\{a_k^l\}_{l=1}^L$.
As \textbf{Corollary \ref{corollary 1}} reveals, we can reduce the generalization error on novel tasks by maximizing the attribute discrimination ability of embedding function $f_\theta$ and the classification ability of prediction function $g_\phi$.
Specifically, we adopt a convolutional network with an additional MLP as $f_\theta$. The convolutional network extracts feature representations from images, then the MLP takes features as input and output attribute labels. 
The attribute classification loss is defined as
\begin{align}
    \mathcal{L}(f_\theta) &= -\frac{1}{m}\sum_{k=1}^m\frac{1}{L}\sum_{l=1}^L \left[a^l_k\log z_k^l + (1-a^l_k)\log (1-z_k^l)\right], 
\end{align}
where $z_k^l$ is the $l$-th dimension of $f_\theta(x_k)$ after a sigmoid function.

For prediction function, we simply choose an non-parametric prediction function like ProtoNet \cite{snell2017prototypical}\footnote{Any other models that map $a\in\mathcal{A}$ to $y\in\mathcal{Y}$ are also feasible, such as a MLP in Relation Network \cite{sung2018relationnet} and a parametric cosine classifier in Baseline++ \cite{chen2019baseline++}.}, which takes the attributes generated by the embedding function $f_\theta$ as input to calculate cosine distance between test samples and attribute prototypes, then predicts the target label.
The few-shot classification loss is defined as
\begin{align}
    \mathcal{F}(f_\theta, g_{\phi_i}) = -\frac{1}{m_i}\sum_{k=1}^{m_i} y_k\log \frac{exp(d_k / t)}{\sum_{y_k}exp(d_k / t)}, 
\end{align}
where $d_k \triangleq cos(f_\theta(x_k), c_{y_k})$ denotes cosine similarity and $c_{y_k}$ denotes the attribute prototype of category $y_k$. 
$t$ is a scalar temperature factor.
In practice, we use a hyperparameter $\beta$ to balance the two losses, so that the final training objective is 
\begin{equation}
    \mathcal{L} = \beta*\mathcal{L}(f_\theta)+\frac{1}{n}\sum_{i=1}^n\mathcal{F}(f_\theta, g_{\phi_i}).
\end{equation}
During the inference phrase, we fix $f_\theta$ then calculate the cosine distance between each query sample and attribute prototypes to predict the target label.

\section{Experiment Details}

\subsection{Implementation Details}

We run experiments with APNet and five classical FSL methods (MatchingNet \cite{vinyals2016matchnet}, ProtoNet \cite{snell2017prototypical}, RelationNet \cite{sung2018relationnet}, MAML \cite{finn2017maml}, Baseline++ \cite{chen2019baseline++}).
Here we explain more implementation details about these methods.
As existing work \cite{chen2019baseline++} has provided a unified testbed for several different FSL methods, we use the codebase and run the experiments for the above methods.
For a fair comparison, we use the four-layer convolution network (Conv4) as backbone model for all methods.
On the CUB dataset, we perform standard data augmentation, including random crop, rotation, horizontal flipping and color jittering, as in \cite{cao2020comet}.
On the SUN dataset, we simply use two augmentation operations, including image scaling and horizontal flipping.
For APNet, we use all provided attribute information (attribute locations and labels) to calculate the attribute classification loss $\mathcal{L}(f_\theta)$.
Because the SUN dataset does not provide attribute locations, we only use attribute labels to calculate $\mathcal{L}(f_\theta)$.
We use the Adam optimizer \cite{kingma2014adam} with an initial learning rate of $10^{-3}$ and weight decay of 0. 
We train models on 5-shot tasks for 40,000 episodes and on 1-shot tasks for 60,000 episodes.
The hyperparameter $\beta$ is tuned on the validation set.
We set $\beta$ to 0.6 and 1.0 for 1-shot and 5-shot setting respectively on CUB dataset, and 0.6 for both settings on SUN dataset.

\subsection{Complete Results}
Here we show complete experimental results which have been partially shown in the main paper.
Tab. \ref{tab:detailed cub and sun} shows the results on CUB and SUN dataset.
Tab. \ref{tab:detailed mini and cross} shows the results on \emph{mini}ImageNet and the cross-dataset scenario (\emph{mini}ImageNet $\rightarrow$ CUB).

\begin{table}[]
\centering
\caption{5-way 1-shot and 5-shot performance of different FSL methods on CUB and SUN datasets. 
Conv4 is used as the backbone model.
We report the average accuracy on 600 novel tasks with 95\% confidence interval.}
\begin{tabular}{lccccc}
\toprule
\multirow{2}{*}{\textbf{Method}} & \multirow{2}{*}{\textbf{Backbone}} & \multicolumn{2}{c}{\textbf{CUB}}                                         & \multicolumn{2}{c}{\textbf{SUN}}                       \\
                        &                           & 5-way 1-shot                         & 5-way 5-shot                         & 5-way 1-shot                & 5-way 5-shot                \\ \midrule
MatchingNet             & \multirow{5}{*}{Conv4}  & 61.02 (0.88)                     & 79.99 (0.75)                     & 57.87 (0.95)            & 76.80 (0.68)            \\
ProtoNet                &                      & 57.12 (0.94)                     & 76.67 (0.65)                     & 60.20 (0.90)           & 76.75 (0.65)            \\
RelationNet             &                      & 61.86 (0.98)                     & 76.63 (0.71)                     & 60.52 (0.91)             & 76.49 (0.65)            \\
MAML                    &                      & 58.73 (0.97)                     & 76.20 (0.69)                     & 59.65 (0.94)            & 76.82 (0.68)            \\
Baseline++              &                      & 60.57 (0.80)                     & 80.17 (0.61)                     & 49.78 (0.82)            & 74.09 (1.11)            \\ \midrule
APNet                   & Conv4                     & 72.96 (0.89) & 85.48 (0.55) & 60.53 (0.86) & 76.35 (0.63) \\ \bottomrule
\end{tabular}
\label{tab:detailed cub and sun}
\end{table}

\begin{table}[]
\centering
\caption{5-way 1-shot and 5-shot performance of different FSL methods on \emph{mini}ImageNet and cross-dataset scenario (\emph{mini}ImageNet$\rightarrow$CUB). 
Conv4 is used as the backbone model.
We report the average accuracy on 600 novel tasks with 95\% confidence interval.}
\begin{tabular}{lccccc}
\toprule
\multirow{2}{*}{\textbf{Method}} & \multirow{2}{*}{\textbf{Backbone}} & \multicolumn{2}{c}{\textbf{\emph{mini}ImageNet}}     & \multicolumn{2}{c}{\textbf{\emph{mini}ImageNet$\rightarrow$CUB}}     \\
                        &                           & 5-way 1-shot       & 5-way 5-shot       & 5-way 1-shot       & 5-way 5-shot       \\ \midrule
MatchingNet             & \multirow{5}{*}{Conv4}    & 49.36 (0.79) & 62.77 (0.69) & 37.48 (0.68) & 49.98 (0.66) \\
ProtoNet                &                           & 42.53 (0.84) & 62.89 (0.72) & 33.91 (0.67) & 53.74 (0.72) \\
RelationNet             &                           & 48.38 (0.80) & 64.37 (0.72) & 38.19 (0.69) & 52.57 (0.66) \\
MAML                    &                           & 45.70 (0.85) & 62.64 (0.72) & 36.97 (0.69) & 51.60 (0.70) \\
Baseline++              &                           & 47.01 (0.71) & 66.72 (0.62) & 37.11 (0.66) & 52.42 (0.67) \\ \bottomrule
\end{tabular}
\label{tab:detailed mini and cross}
\end{table}


{{
      \bibliographystyle{IEEEtran}
      \bibliography{overleaf/bibtex_v2310/egbib, overleaf/bibtex_v2310/bilevel_optimization, overleaf/bibtex_v2310/domain_adaptation, overleaf/bibtex_v2310/few-shot_learning, overleaf/bibtex_v2310/graph_neural_networks, overleaf/bibtex_v2310/others, overleaf/bibtex_v2310/transfer_learning, overleaf/bibtex_v2310/scene_graph, overleaf/bibtex_v2310/theory}

}}